\def\paperID{486}
\def\confName{CVPR}
\def\confYear{2024}

\def\paperTitle{SHINOBI: \emph{\textcolor{darkred}{Sh}}ape and \emph{\textcolor{darkred}{I}}llumination using \emph{\textcolor{darkred}{N}}eural \emph{\textcolor{darkred}{O}}bject Decomposition via \\ \emph{\textcolor{darkred}{B}}RDF Optimization \emph{\textcolor{darkred}{I}}n-the-wild}

\def\authorBlock{
Andreas Engelhardt$^\dag$ \\
University of T{\"{u}}bingen \\
\and
Amit Raj \\
Google Research \\
\and
Mark Boss$^*$ \\
Unity \\
\and
Yunzhi Zhang$^\dag$ \\
Stanford University \\
\and
Abhishek Kar \\
Google Research \\
\and
Yuanzhen Li \\
Google Research \\
\and
Deqing Sun \\
Google Research \\
\and
Ricardo Martin Brualla \\
Google Research \\
\and
Jonathan T. Barron \\
Google Research \\
\and
Hendrik P. A. Lensch \\
University of T{\"{u}}bingen \\%
\and
Varun Jampani$^*$ \\
Google Research \\

}

\newif\ifreview 
\newif\ifarxiv \newcommand{\arxiv}{\arxivtrue}
\newif\ifcamera 
\newif\ifrebuttal 

\newif\ifdrafting
\draftingtrue %
\ifdrafting
\newcommand{\aen}[1]{{\color{blueish} [AE: #1]}} %
\newcommand{\mb}[1]{{\color{blueish}#1}} %
\newcommand{\MB}[1]{{\color{blueish}{\bf [MB: #1]}}} %
\newcommand{\Mb}[1]{\marginpar{\tiny{\textcolor{blueish}{#1}}}} %
\newcommand{\vj}[1]{{\color{dark_green}{\bf [VJ: #1]}}}
\newcommand{\VJ}[1]{{\color{dark_green}{\bf [VJ: #1]}}}
\newcommand{\Vj}[1]{\marginpar{\tiny{\textcolor{dark_green}{#1}}}}
\newcommand{\jb}[1]{{\color{greyblue}#1}}
\newcommand{\JB}[1]{{\color{greyblue}{\bf [JB: #1]}}}
\newcommand{\Jb}[1]{\marginpar{\tiny{\textcolor{greyblue}{#1}}}}
\newcommand{\ar}[1]{{\color{mgreen}#1}}
\newcommand{\AR}[1]{{\color{mgreen}{\bf [AR: #1]}}}
\newcommand{\Ar}[1]{\marginpar{\tiny{\textcolor{mgreen}{#1}}}}
\newcommand{\todo}[1]{{\color{red}#1}}
\newcommand{\TODO}[1]{\textbf{\color{red}[TODO: #1]}}
\newcommand{\hl}[1]{{\color{brown}{\bf [HL: #1]}}}

\newcommand{\ds}[1]{{\color{red}[DS: #1]}}
\else
\newcommand{\aen}[1]{}
\newcommand{\mb}[1]{} 
\newcommand{\MB}[1]{} 
\newcommand{\Mb}[1]{} 
\newcommand{\vj}[1]{}
\newcommand{\VJ}[1]{}
\newcommand{\Vj}[1]{}
\newcommand{\jb}[1]{}
\newcommand{\JB}[1]{}
\newcommand{\Jb}[1]{}
\newcommand{\ar}[1]{}
\newcommand{\AR}[1]{}
\newcommand{\Ar}[1]{}
\newcommand{\todo}[1]{}
\newcommand{\TODO}[1]{}
\newcommand{\hl}[1]{}
\newcommand{\ds}[1]{}
\fi

\arxiv %

\pdfoutput=1
\documentclass[10pt,twocolumn,letterpaper]{article}
\ifreview \usepackage[review]{cvpr} \fi
\ifarxiv \usepackage[pagenumbers]{cvpr} \fi
\ifrebuttal \usepackage[rebuttal]{cvpr} \fi
\ifcamera \usepackage{cvpr} \fi

\usepackage{graphicx}	
\usepackage{amsmath}	
\usepackage{amssymb}	
\usepackage{booktabs}
\usepackage{colortbl}
\usepackage{times}
\usepackage{microtype}
\usepackage{epsfig}
\usepackage[table,xcdraw,dvipsnames]{xcolor}
\usepackage{caption}
\usepackage{float}
\usepackage{placeins}
\usepackage{color, colortbl}
\usepackage{stfloats}
\usepackage{enumitem}
\usepackage{tabularx}
\usepackage{xstring}
\usepackage{multirow}
\usepackage{xspace}
\usepackage{xfrac}
\usepackage{url}
\usepackage{subcaption}
\usepackage{xcolor}
\usepackage{colortbl}
\usepackage{bm}
\usepackage[hang,flushmargin]{footmisc}

\usepackage{siunitx}

\ifcamera \usepackage[accsupp]{axessibility} \fi

\usepackage{tikz}

\usetikzlibrary{shapes}
\usetikzlibrary{calc}
\usetikzlibrary{backgrounds}
\usetikzlibrary{positioning}
\usetikzlibrary{arrows}
\usetikzlibrary{arrows.meta}
\usetikzlibrary{decorations.text}    
\usetikzlibrary{fit}
\usetikzlibrary{spy}
\usetikzlibrary{3d}
\usetikzlibrary{math}

\definecolor{turquoise}{cmyk}{0.65,0,0.1,0.3}
\definecolor{purple}{rgb}{0.65,0,0.65}
\definecolor{dark_green}{rgb}{0, 0.5, 0}
\definecolor{orange}{rgb}{0.8, 0.6, 0.2}
\definecolor{red}{rgb}{0.8, 0.2, 0.2}
\definecolor{darkred}{rgb}{0.6, 0.1, 0.05}
\definecolor{blueish}{rgb}{0.0, 0.3, .6}
\definecolor{light_gray}{rgb}{0.7, 0.7, .7}
\definecolor{pink}{rgb}{1, 0, 1}
\definecolor{greyblue}{rgb}{0.25, 0.25, 1}

\definecolor{bestcol}{RGB}{254,196,79}
\newcommand{\best}[1]{\cellcolor{bestcol} \textbf{#1}}
\definecolor{secondbestcol}{RGB}{255,247,188}
\newcommand{\secondbest}[1]{\cellcolor{secondbestcol} #1}

\newcommand{\OURS}{SHINOBI~}

\ifarxiv  \fi

\newcommand{\R}[1]{{%
    \textbf{%
        \ifstrequal{#1}{1}{\textcolor{red}{R1}}{%
        \ifstrequal{#1}{2}{\textcolor{blueish}{R2}}{%
        \ifstrequal{#1}{3}{\textcolor{purple}{R3}}{%
        \ifstrequal{#1}{4}{\textcolor{teal}{R#1}}{%
                           \textcolor{cyan}{R#1}%
        }}}}%
    }%
}}

\DeclareMathOperator{\arctantwo}{arctan2}
\newcommand{\loss}[1]{\mathcal{L}_\mathrm{#1}}
\newcommand{\losssuperscripted}[2]{\mathcal{L}_{\mathrm{#1}}^{(#2)}}

\newcommand{\vect}[1]{\bm{#1}}

\newcommand{\expnumber}[2]{{#1}e{#2}}

\newcommand\lft{\mathopen{}\left}
\newcommand\rgt{\aftergroup\mathclose\aftergroup{\aftergroup}\right}

\newcommand{\maxclip}[2]{\ensuremath{\operatorname{max}\lft(#1, #2\rgt)}}

\newcommand{\fig}[1]{Fig.~\ref{fig:#1}}

\makeatletter
\DeclareRobustCommand\onedot{\futurelet\@let@token\@onedot}
\def\@onedot{\ifx\@let@token.\else.\null\fi\xspace}

\def\eg{\emph{e.g}\onedot}

\def\vs{\emph{vs}\onedot}
\def\wrt{w.r.t\onedot} 

\def\etal{\emph{et al}\onedot}

\makeatother

\newcommand{\inlinesection}[1]{\vspace{0.05cm} \noindent {\bf #1}}
\newcommand{\titlecaption}[2]{\caption{\textbf{#1.}\xspace#2}}

\newcommand{\titlecaptionof}[3]{\captionof{#1}{\textbf{#2.}\xspace#3}}

\usepackage{pifont}

\usepackage{blindtext}

\definecolor{mpurple}{RGB}{106,27,154}
\definecolor{mpurplelight}{RGB}{206,147,216}

\definecolor{mblue}{RGB}{40,53,147}
\definecolor{mbluelight}{RGB}{159,168,218}

\definecolor{mteal}{RGB}{0,105,92}
\definecolor{mteallight}{RGB}{128,203,196}

\definecolor{morangelight}{RGB}{255,171,145}

\definecolor{mgrayblue}{RGB}{55,71,79}
\definecolor{mgraybluelight}{RGB}{176,190,197}

\definecolor{mamber}{RGB}{255,143,0}
\definecolor{mamberlight}{RGB}{255,224,130}

\definecolor{mdeeporange}{RGB}{216,67,21}
\definecolor{morange}{RGB}{245,124,0}
\definecolor{myellow}{RGB}{253,216,53}

\definecolor{mgreen}{RGB}{85,139,47}
\definecolor{mgreenlight}{RGB}{174,213,129}

\definecolor{mred}{RGB}{198,40,40}
\definecolor{mredlight}{RGB}{239,154,154}

\definecolor{corange1}{RGB}{254,237,222}
\definecolor{corange2}{RGB}{253,190,133}
\definecolor{corange3}{RGB}{253,141,60}
\definecolor{corange4}{RGB}{230,85,13}
\definecolor{corange5}{RGB}{166,54,3}

\definecolor{cblue1}{RGB}{239,243,255}
\definecolor{cblue2}{RGB}{189,215,231}
\definecolor{cblue3}{RGB}{107,174,214}
\definecolor{cblue4}{RGB}{49,130,189}
\definecolor{cblue5}{RGB}{8,81,156}

\definecolor{cgray1}{RGB}{247,247,247}
\definecolor{cgray2}{RGB}{204,204,204}
\definecolor{cgray3}{RGB}{150,150,150}
\definecolor{cgray4}{RGB}{99,99,99}
\definecolor{cgray5}{RGB}{37,37,37}

\definecolor{cred1}{RGB}{254,229,217}
\definecolor{cred2}{RGB}{252,174,145}
\definecolor{cred3}{RGB}{251,106,74}
\definecolor{cred4}{RGB}{222,45,38}
\definecolor{cred5}{RGB}{165,15,21}

\definecolor{cgreen1}{RGB}{237,248,233}
\definecolor{cgreen2}{RGB}{186,228,179}
\definecolor{cgreen3}{RGB}{116,196,118}
\definecolor{cgreen4}{RGB}{49,163,84}
\definecolor{cgreen5}{RGB}{0,109,44}

\definecolor{cdiv11}{RGB}{166,97,26}
\definecolor{cdiv12}{RGB}{223,194,125}
\definecolor{cdiv13}{RGB}{245,245,245}
\definecolor{cdiv14}{RGB}{128,205,193}
\definecolor{cdiv15}{RGB}{1,133,113}

\definecolor{cdiv21}{RGB}{208,28,139}
\definecolor{cdiv22}{RGB}{241,182,218}
\definecolor{cdiv23}{RGB}{247,247,247}
\definecolor{cdiv24}{RGB}{184,225,134}
\definecolor{cdiv25}{RGB}{77,172,38}

\definecolor{cdiv31}{RGB}{230,97,1}
\definecolor{cdiv32}{RGB}{253,184,99}
\definecolor{cdiv33}{RGB}{247,247,247}
\definecolor{cdiv34}{RGB}{178,171,210}
\definecolor{cdiv35}{RGB}{94,60,153}

\definecolor{cdiv41}{RGB}{216,179,101}
\definecolor{cdiv42}{RGB}{245,245,245}
\definecolor{cdiv43}{RGB}{90,180,172}

\usepackage{xr-hyper}

\makeatletter
\newcommand*{\addFileDependency}[1]{
  \typeout{(#1)}
  \@addtofilelist{#1}
  \IfFileExists{#1}{}{\typeout{No file #1.}}
}

\makeatother

\definecolor{cvprblue}{rgb}{0.21,0.49,0.74}
\usepackage[pagebackref,breaklinks,colorlinks,citecolor=cvprblue]{hyperref}
\usepackage[capitalize]{cleveref}
\crefname{section}{Sec.}{Secs.}
\crefname{table}{Table}{Tables}
\crefname{figure}{Fig.}{Figs.}

\title{\paperTitle}

\author{\authorBlock} 
\begin{document}

\twocolumn[{%
            \renewcommand\twocolumn[1][]{#1}%
            \maketitle

\begin{center}
    \centering
    \captionsetup{type=figure}
    \includegraphics[trim=0 520 0 0,clip,width=\textwidth]{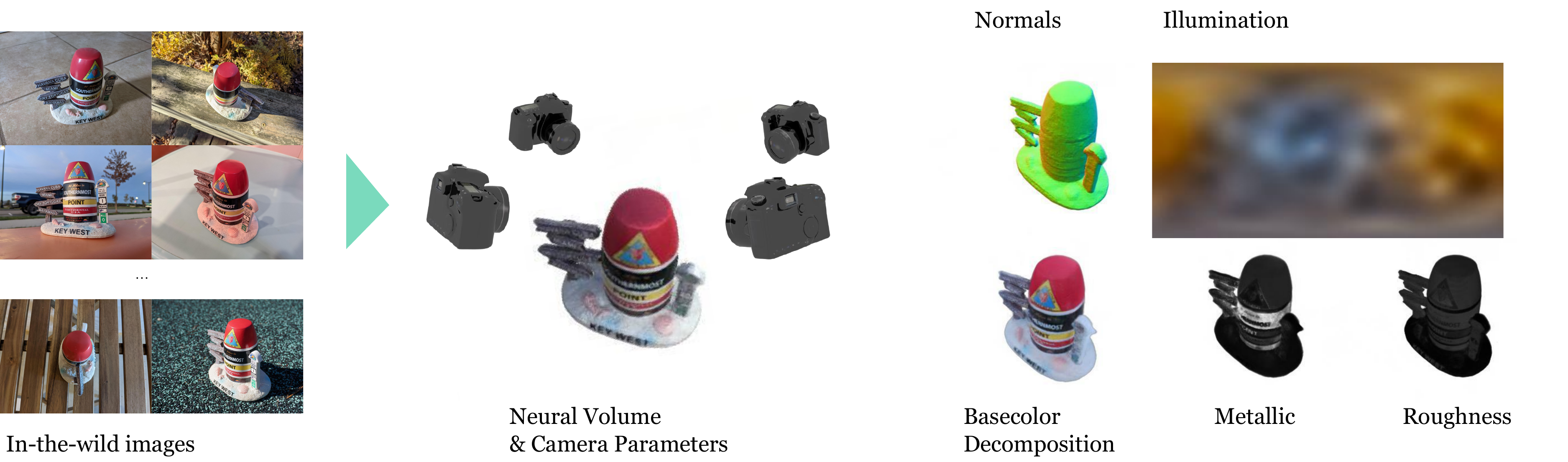}
    \titlecaption{Object reconstruction using SHINOBI}{SHINOBI decomposes challenging in-the-wild image collections 
into shape, material and illumination using a neural field representation while also optimizing camera parameters. Visit the project page at Project page: \url{https://shinobi.aengelhardt.com}} %
\label{fig:teaser}
\end{center}%

        }]

\renewcommand{\thefootnote}{\fnsymbol{footnote}}
\footnotetext[1]{Current affiliation is Stability AI.}
\footnotetext[2]{Work done during a Student Researcher position at Google.}
\begin{abstract}
We present SHINOBI, an end-to-end framework for the reconstruction of shape, material, and illumination from object images captured with varying lighting, pose, and background. Inverse rendering of an object based on unconstrained image collections is a long-standing challenge in computer vision and graphics and requires a joint optimization over shape, radiance, and pose. 
We show that an implicit shape representation based on a multi-resolution hash encoding enables faster and robust shape reconstruction with joint camera alignment optimization that outperforms prior work. Further, to enable the editing of illumination and object reflectance (i.e. material) we jointly optimize BRDF and illumination together with the object's shape. %
Our method is class-agnostic and works on in-the-wild image collections of objects to produce relightable 3D assets for several use cases such as AR/VR, movies, games, etc.
\end{abstract}

\vspace{-3mm}
\section{Introduction}
\label{sec:intro}
\vspace{-2mm}

\renewcommand{\thefootnote}{\fnsymbol{footnote}}

We present a category-agnostic technique to jointly reconstruct 3D shape and material properties of objects from unconstrained in-the-wild image collections. 
This data regime poses multiple challenges as images are captured in different environments using a variety of devices resulting in varying backgrounds, illumination, camera poses, and intrinsics. In addition, camera baselines tend to be large.  \fig{teaser} (left) shows examples from an input image set. 
Many graphics applications in AR/VR, games, and movies depend on high-quality 3D assets of real-world objects. Physically based materials are essential to integrate objects into new environments. The conventional acquisition involves laborious tasks like 3D modeling, texture painting, and light calibration or use controlled setups~\cite{bi2020a,Nam2018} that are hard to scale.
It is easier to obtain casually captured images from smartphones or image collections from the internet for a large number of objects.

Conventional structure-from-motion techniques like COLMAP~\cite{schoenberger2016mvs,schoenberger2016sfm} fail to reconstruct image collections under these challenging circumstances~\cite{bossSAMURAIShapeMaterial2022, jampani2023navi}. Despite constraining the correspondences to lie within object bounds, specifically in the context of the NAVI~\cite{jampani2023navi} in-the-wild scenes, less than half of the views are registered on average with half the scenes failing completely. 
Consequently, we observe that camera pose optimization has the largest impact on the reconstruction quality in this setting. 
Many existing works on shape and material estimation~\cite{Boss2021neuralPIL,bi2020a,srinivasan2020,zhang2021,zhang2022invrender,neilf2022Yao,Ye2023IntrinsicNeRF} assume constant camera intrinsics and initialization of camera poses close to the true poses.
We support 360\textdegree~multiview data with a rough quadrant-based pose initialization with poses potentially far from the ground truth,
as in SAMURAI~\cite{bossSAMURAIShapeMaterial2022} and NeRS~\cite{zhang2021ners}. For challenging data this can be annotated in only a few minutes per image collection.
Even though in SAMURAI~\cite{bossSAMURAIShapeMaterial2022}, camera poses can be initialized from very coarse directions slight offsets often lead to overly smooth textures and shapes in the final reconstructions.
Further, existing methods for material decomposition with camera pose optimization are slow, often running more than 12 hours on a single object~\cite{bossSAMURAIShapeMaterial2022,kuang2022neroic}. 
In contrast, we propose a pipeline based on multiresolution hash grids~\cite{Mueller2022} which allows us to process more rays in a shorter time during optimization. Using this advantage we are able to improve reconstruction quality compared to SAMURAI while still keeping a competitive run-time (Tab.~\ref{tab:3d_wild_synthesis_gt}).

Naive integration of multi-resolution hash grids is not well suited to camera pose estimation due to discontinuities in the gradients with respect to the input positions. We propose several components that work together to stabilize the camera pose optimization and encourage sharp features. The key distinguishing features of \OURS include:

\begin{itemize}[leftmargin=*, itemsep=0mm] %
\item \textit{Hybrid Multiresolution Hash Encoding with level annealing.}
We combine the multiresolution hash-based encoding~\cite{Mueller2022} with regular Fourier feature transformation of the input coordinates to regularize the low-frequency gradient propagation. This makes the optimization significantly more robust while only adding a small overhead.
A similar approach has been recently proposed by Zhu~\etal~\cite{zhu2023rhino} for a different task. We show that it is also beneficial for camera pose optimization. 

\item \textit{Camera multiplex constraint.}
We modify the camera parameterization of SAMURAI to avoid over-parameterization of the camera rotations. 
Furthermore, we constrain the camera optimization with a projection-based loss to enforce consistency over the camera proposals inside a multiplex which further helps to smooth the optimization in the initial phase.

\item \textit{Per-view importance weighting.} 
We propose a per-view importance weighting to leverage the important observation that some views are more useful for optimization than others. Specifically, we use well-working cameras to anchor the reconstruction during the optimization.
\item \textit{Patch-based alignment losses.} 
SHINOBI proposes a novel patch level loss to aid in camera alignment and additionally introduces a silhouette loss inspired by Lensch~\etal~\cite{automatedLensch2000} for better image to 3D alignment.
\end{itemize}

Experiments on NAVI~\cite{jampani2023navi} in-the-wild datasets demonstrate better view synthesis and relighting results with~\OURS compared to existing works with a reduced run-time. Compared to SAMURAI the results look sharper and the average runtime is cut in half. \fig{teaser} (right) shows some sample application results with 3D assets generated by~\OURS. 
Our representation enables editing of appearance parameters, illumination and based on the mesh extraction also shape, facilitating various tasks in a downstream graphics pipeline.

\begin{figure*}[t]
\begin{center}
\begin{tikzpicture}[
    node distance=0.5cm, 
    font={\fontsize{8pt}{10}\selectfont},
    layer/.style={draw=#1, fill=#1!20, line width=1.5pt, inner sep=0.2cm, rounded corners=1pt},
    textlayer/.style={draw=#1, fill=#1!20, line width=1.5pt, inner sep=0.1cm, rounded corners=1pt},
    encoder/.style={isosceles triangle,isosceles triangle apex angle=60,shape border rotate=0,anchor=apex,draw=#1, fill=#1!20, line width=1.5pt, inner sep=0.1cm, rounded corners=1pt},
    decoder/.style={isosceles triangle,isosceles triangle apex angle=60,shape border rotate=180,anchor=apex,draw=#1, fill=#1!20, line width=1.5pt, inner sep=0.1cm, rounded corners=1pt},
    label/.style={font=\scriptsize, text width=1.2cm, align=center},
    fourier/.pic={%
        \node (#1){
            \begin{tikzpicture}[remember picture]%
                \draw[line width=0.75pt, fill=cdiv42] (0,0) circle (6pt);%
                \draw[line width=0.75pt] (-6pt,0.0) sin (-3pt,2pt) cos (0,0) sin (3pt,-2pt) cos (6pt,0);%
            \end{tikzpicture}
        };
    },
    hashgrid/.pic={%
        \node (#1) {
            \begin{tikzpicture}[remember picture]%
                \draw[line width=0.75pt, draw=cdiv43, fill=cdiv43!20] (0.2, 0.4) rectangle (1.2, 1.2);
                \draw[line width=0.75pt, draw=cdiv43, fill=cdiv43!20] (0.1, 0.3) rectangle (1.1, 1.1);
                \draw[line width=0.75pt, fill=cdiv43!20] (0, 0) rectangle (1, 1);
                \draw[step=0.2,black!70,thin] (0, 0) grid (1,1);
                \draw[line width=0.75pt, draw=cdiv43] (0, 0) rectangle (1, 1);
            \end{tikzpicture}
        };
    },
    annealing/.pic={%
        \node (#1) {
            \begin{tikzpicture}[remember picture]%
            \begin{scope} [rotate=180]
                    \fill[fill=cdiv42] (0, 0) circle (0.2);
                    \draw[draw=black, fill=cdiv42]
                    \foreach \i in {1,2,...,10} {%
                       [rotate=(\i-1)*36]  (0:0.2)  arc (0:12:0.2) -- (18:{0.2+max(cos((\i-5)*25),0)*0.05}) arc (18:30:{0.2+max(cos((\i-5) *25),0)*0.05}) -- (36:0.2) 
                     }; %
            \end{scope}
            \end{tikzpicture}
        };
    }
]

\node[] (main_img) at (0.15, 0.3) {\includegraphics[width=3.6cm] {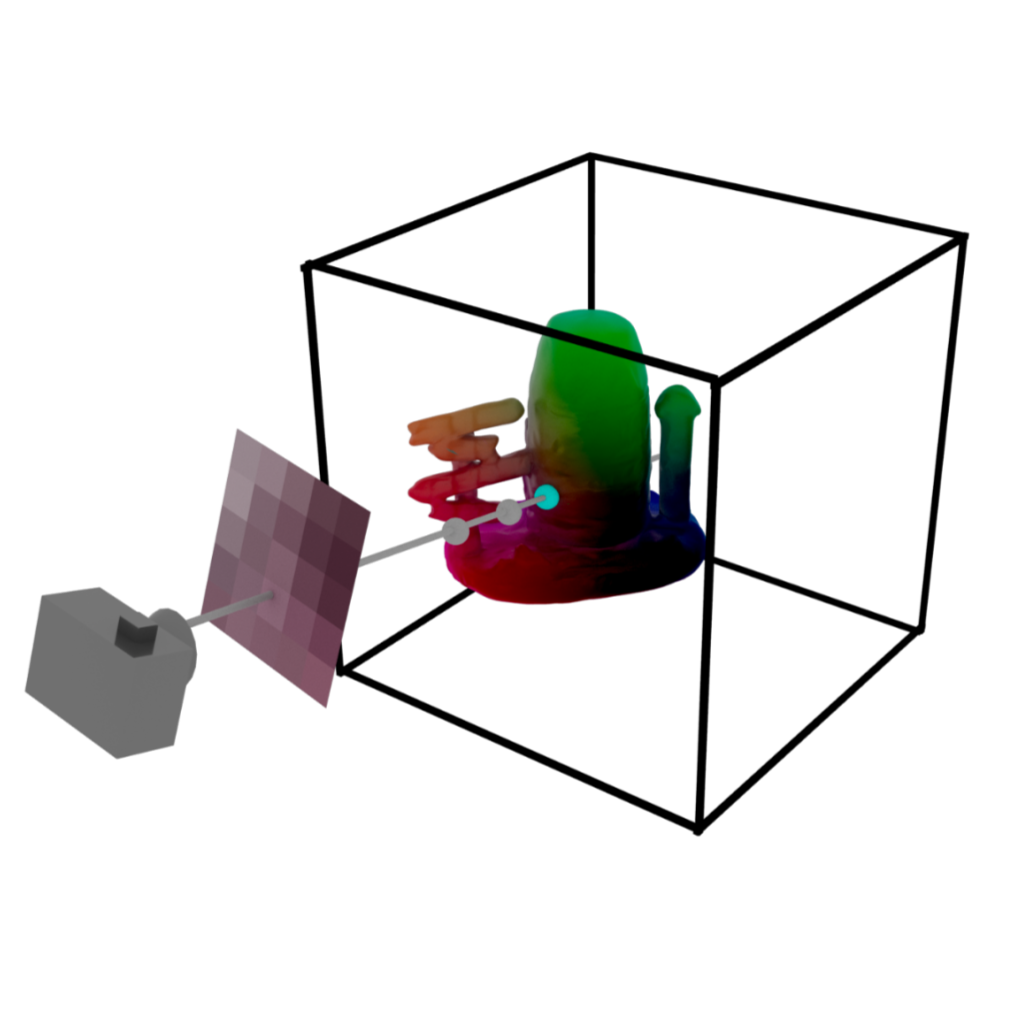}};
\node(gt_img) at (-0.8, 2.2){\frame{\includegraphics[width=1.5cm,trim=100 20 20 10,clip]{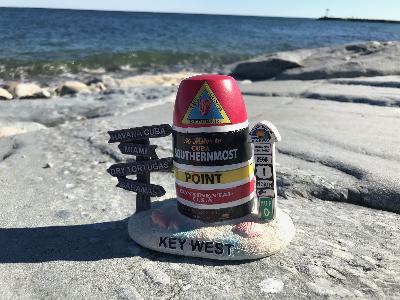}}};
\draw[thick] ($(gt_img.south)+(0.0,0.3)$) rectangle ($(gt_img.south)+(0.4,0.7)$);

\node[] (positions) at (2, -1.5) {Positions $\vect{x}_t$};
\node[] (pixel) at (-0.8, -0.6) {$C_j^s$};
\node[] (patch_gt) at (-1.0, 0.65) {$P_j^s$}; %
\draw[fill=black] (-0.68,0.04) circle (1.2pt);
\draw[-stealth, dashed] ($(gt_img.south)+(0.0,0.3)$) to[out=200, in=90] (patch_gt);

\node[text width=0.7cm, align=center] (direction) at (-0.9, -1.75) {Direction $\vect{d}$};

\draw[-stealth, dashed, draw=cgreen5] (-0.03, 0.2) to[out=340, in=90] (positions);
\draw[-stealth, dashed, draw=cgreen5] (0.175, 0.25) to[out=340, in=90] (positions);
\draw[-stealth, dashed, draw=cgreen5] (0.36, 0.33) to[out=340, in=90] (positions);

\node[textlayer={cdiv43}, text width=1.5cm, align=center] (intrinsics) at (-3, 0.5) {Intrinsics $\hat{f}^j$};
\node[textlayer={cdiv43}, text width=1.6cm, align=center, below=0.1cm of intrinsics] (extrinsics) {Extrinsics $\vect{p}_\text{eye}^j, \vect{d}_{\phi\theta}^j, d_\text{up}^j$};

\pic at (3.25, 0.8) {hashgrid=hashgrid};
\node[text width=0.4cm, align=center, above=-0.2cm of hashgrid] (hashgrid_label) {$H(\vect{x})$};
\pic at (2.9, -0.6) {fourier=fourier};
\node[textlayer={cdiv43}, font=\tiny, right=0.15cm of fourier] (embd_mlp) {\rotatebox{90}{MLP}};
\node[text width=0.4cm, align=center, below=0.225cm of $(embd_mlp)!0.5!(fourier)$] (fourier_label) {$\gamma(\vect{x})$};
\pic at (4.5, 0.8) {annealing=annealing};

\draw[-stealth, shorten <=-0.1cm, shorten >=0.075cm] (fourier) -- (embd_mlp);
\draw[-stealth, shorten >=-0.1cm] (positions) to[out=90, in=240] (fourier);
\draw[-stealth, shorten >=-0.1cm] (positions) to[out=90, in=240] (hashgrid);

\coordinate (between_embds) at ($(embd_mlp)!0.5!(hashgrid)$);

\draw (embd_mlp.east) -| ($(between_embds)+(0.6,0.0)$);
\draw[shorten <=-0.15cm] (hashgrid.east) -| ($(between_embds)+(0.6,0.0)$);

\node[text width=1cm, align=center, above=-0.2cm of annealing.north] (annealing_label) {Annealing};
\coordinate (anneal_join) at ($(between_embds)+(1.0,0.0)$);
\draw[shorten <=-0.13cm] (annealing.south) |- (anneal_join);
\draw ($(between_embds)+(0.6,0.0)$) |- (anneal_join);

\begin{scope}[on background layer]
    \draw[layer={cdiv41}, rounded corners=1pt,line width=1.5pt] ($(hashgrid.north west)+(-0.075,0.275)$)  rectangle ($(embd_mlp.south east)+(1.675,-0.375)$);
\end{scope}

\node[text width=5cm, align=center, above=1cm of $(hashgrid)!0.5!(annealing)$] (encoding_box_label) {Annealed Hybrid Encoding};

\node[textlayer={cdiv42}, font=\tiny, right=1.5cm of anneal_join] (main_mlp1) {\rotatebox{90}{MLP}};
\node[textlayer={cdiv42}, font=\tiny, right=0.075cm of main_mlp1] (main_mlp2) {\rotatebox{90}{MLP}};

\begin{scope}[on background layer]
    \draw[layer={cdiv43}, rounded corners=1pt,line width=1.5pt] ($(main_mlp1.north west)+(-0.075,0.075)$)  rectangle ($(main_mlp2.south east)+(0.075,-0.075)$);
\end{scope}
\node[above=0.1cm of $(main_mlp1.north)!0.5!(main_mlp2.north)$, label, text width=1.5cm] {Network};

\node[label, text width=0.9cm, right=0.5cm of main_mlp2] (brdf) {BRDFs $\vect{b}_t$};
\node[label, text width=0.9cm, below=0.3cm of brdf] (density) {Densities $\sigma_t$};

\node[textlayer={cdiv43}, text width=1.5cm, align=center, above=0.6cm of brdf] (illumination) {Illumination $\vect{z}_j$};
\node[textlayer={gray}, right=0.25cm of brdf] (int) {$\int$};
\node[textlayer={cdiv41}, right=0.25cm of int] (renderer) {\rotatebox{90}{\tiny PIL-Renderer}};

\node[label, text width=0.9cm, right=0.5cm of renderer] (color) {Color $\vect{\hat{c}}_j$};

\node[above=of color] (pred_img) {\frame{\includegraphics[width=1.5cm,trim=100 20 20 10,clip]{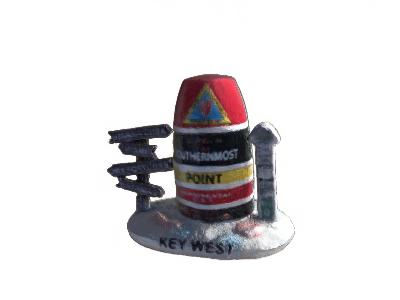}}};
\draw[thick] ($(pred_img.south)+(0.0,0.3)$) rectangle ($(pred_img.south)+(0.4,0.7)$);

\draw[dashed] (color) to[out=90, in=270] ($(pred_img.south)+(0.2,0.3)$);

\draw[dashed] ($(gt_img.south)+(0.4,0.7)$) to[out=20, in=160] ($(pred_img.south)+(0.0,0.7)$);
\node at (5,2.75) {Patch-based Losses};

\draw[-stealth, shorten <=-0.1cm, shorten >=0.075cm] (anneal_join) -- (main_mlp1);
\draw[-stealth, shorten <=0.1cm] (main_mlp2) -- (brdf);
\draw[-stealth] ($(main_mlp2)!0.4!(brdf)$) |- (density);
\draw[-stealth] (brdf) -- (int);
\draw[-stealth] (density) -| (int);
\draw[-stealth] (int) -- (renderer);
\draw[-stealth] (illumination) -| (renderer);
\draw[-stealth] (direction) -| (renderer);
\draw[-stealth] (renderer) -- (color);

\node[textlayer={cdiv43}, text width=0.25cm, text height=0.25cm] at (-3.6, 1.75) (icon_learnable) {};
\node[label, right=0.05cm of icon_learnable.east, align=left] {\textbf{:}};
\node[label, right=0.3cm of icon_learnable.east, align=left] {Optimizable Prameters};

\end{tikzpicture}
\vspace{-4mm}
\end{center}
\titlecaption{The SHINOBI pipeline}{Two resolution annealed encoding branches, the multiresolution hash grid $H(\vect{x})$ and the Fourier embedding $\gamma(\vect{x})$ are used to learn a neural volume conditioned on the input coordinates. This enables robust optimization of camera parameters jointly with the shape, material and illumination.}
\label{fig:overview}
\end{figure*}

\vspace{-2mm}
\section{Related works}
\label{sec:related}
\vspace{-2mm}

\noindent{\bf Neural fields} have emerged as a popular technique of late to encode spatial information in the network weights of \eg an MLP, which can be retrieved by simply querying the coordinates~\cite{chen2018, mescheder2019occupancy, DeepSDFPark2019, tancik2020fourfeat}. 
Works like NeRF~\cite{mildenhall2020} leverage this neural volume rendering to achieve photo-realistic view synthesis results with view-dependent appearance variations.
Rapid research in neural fields followed, which alternated the surface representations~\cite{Wang2021neus, Oechsle2021, volsdfyariv2021, neus2, neddfUeda2022, xuHybridMeshneuralRepresentation2022},
allowed reconstruction from sparse data~\cite{Niemeyer2021Regnerf, sparf2023Truong, rebain2022lolnerf, long2022sparseneus, dietnerfJain2021,Yang2023FreeNeRF, bian2022nopenerf}, enabled extraction of 3D geometry and materials~\cite{Boss2021, zhang2021ners, munkberg2022nvdiffrec, kuang2022neroic}, 
or enabled relighting of scenes \cite{bi2020b, Boss2021, Boss2021neuralPIL, bossSAMURAIShapeMaterial2022, martinbrualla2020nerfw, liangENVIDRImplicitDifferentiable2023, Ye2023IntrinsicNeRF}. 
However, most prior works rely on pose information extracted from COLMAP ~\cite{schoenberger2016mvs, schoenberger2016sfm}, which can be inaccurate or completely fail in complex settings or sparse data regimes.
\OURS is independent of any \emph{pose reconstruction that relies on feature matching} and robust to very \emph{coarse initialization}. 

\noindent{\bf Instant Neural Graphics Primitives}  (I-NGP)~\cite{Mueller2022} is a popular geometric representation that enables fast optimization with improved memory utilization by using an encoding scheme based on multi-resolution hash tables.
Despite the improvement in speed, I-NGP suffers from discontinuous and oscillating gradient flow through the hash-based encoding, which complicates camera pose optimization~\cite{zhu2023rhino, robustPoseHeo2023, zhu2023rhino}. To enable reconstruction with camera pose fine-tuning using hash grids, Heo \etal~\cite{robustPoseHeo2023} propose a modification to the interpolation weighting, BAA-NGP~\cite{liu2023baangp} dynamically replicates low-resolution features and CAMP~\cite{park2023camp} pairs a robust sampling scheme~\cite{barron2023zipnerf} with camera preconditioning. These methods however are sensitive to camera initialization and lighting conditions.
In contrast to these works, SHINOBI is able to %
reconstruct consistent objects from images captured under \emph{varying illuminations and backgrounds} besides supporting \emph{coarser poses.}

\noindent{\bf Joint camera and shape estimation} is a highly ambiguous task, traditionally relying on accurate poses for precise shape reconstruction and vice versa.
Often techniques rely on correspondences across images to estimate camera poses~\cite{schoenberger2016mvs, schoenberger2016sfm}.
Recent approaches integrate camera calibration with neural volume training; SCNeRF~\cite{SCNeRF2021} and NopeNeRF~\cite{bian2022nopenerf} use correspondences and monocular depth images, respectively.
Other recent methods rely on rough initialization of the camera, global alignment, or a template shape for joint optimization \cite{lin2021barf,zhang2021ners,wang2021nerfmm,chen2023local2global}. Other methods use transformer-based models\cite{ViTDosovitskiyB0WZ21} to predict the initial pose from image collection \cite{sparseposeSinha2023,Zhang2022VMRF}.
In comparison, \OURS works on \emph{unconstrained image collections}, including various camera parameters and object environments, where existing methods struggle to generalize or require additional input data like depth.

\noindent{\bf BRDF and illumination estimation} is a challenging and ambiguous problem.
Casual BRDF estimation enables on-site material acquisition with simple cameras and a co-located camera flash.
These techniques often constrain the problem to planar surfaces with either a single shot~\cite{Aittala2018, Deschaintre2018, henzler2021, Li2018, sang2020single, Boss2019}, few-shot~\cite{Aittala2018} or multi-shot~\cite{Albert2018, Boss2018, Deschaintre2019, Deschaintre2020,Gao2019} captures.
Casual capture can also be extended to a joint BRDF and shape reconstruction~\cite{bi2020a,bi2020b,bi2020c,Boss2020,kaya2021uncalibrated,Nam2018,sang2020single,Zhang2020InverseRendering}, even on entire scenes~\cite{li2020inverse,Sengupta2019}.
Most of these methods, however, require a known active illumination.
Recovering a BRDF under unknown passive illumination is significantly more challenging as it requires disentangling the BRDF from the illumination.
Recently, neural field-based decomposition achieved decomposition of scenes under varying illumination~\cite{Boss2021, Boss2021neuralPIL} or fixed illumination~\cite{zhang2021, liangENVIDRImplicitDifferentiable2023,zhang2023neilf++, nerfactor_zhang21,zhang2022invrender}. IntrinsicNeRF~\cite{Ye2023IntrinsicNeRF} extends decomposition to larger scenes at the cost of a simplified reflectance model. However, all these approaches require known, near-perfect camera poses, whereas SHINOBI can work with \emph{unposed image collection} to recover per-image illumination.

\vspace{-2mm}
\section{Method}
\label{sec:method}
\vspace{-2mm}

The aim of \OURS is to convert 2D image collections into a 3D representation with minimal manual work. The representation includes shape, material parameters and per-view illumination, allowing for view synthesis with relighting. 

\inlinesection{Problem setup.}
We define in-the-wild data as a collection of $q$ images $C_j \in \mathbb{R}^{s_j \times 3}; j \in \{1,\ldots,q\}$ that show the same object captured with different backgrounds, illuminations and cameras with potentially varying resolutions $s_j$. 
In addition, we assume a rough camera initialization. For our experiments we annotate camera pose quadrants
as in SAMURAI~\cite{bossSAMURAIShapeMaterial2022}. 
Foreground masks can be added if available or automatically generated 
and might be imperfect at this point.
At each point $\vect{x} \in \mathbb{R}^{3}$ in the neural volume $\mathcal{V}$, we estimate the BRDF parameters for the Cook-Torrance model~\cite{Cook1982} $\vect{b} \in \mathbb{R}^{5}$ (basecolor $\vect{b}_c \in \mathbb{R}^{3}$, metallic $b_m \in \mathbb{R}$, roughness $b_r \in \mathbb{R}$), unit-length surface normal $\vect{n} \in \mathbb{R}^{3}$ and volume density $\sigma \in \mathbb{R}$ (Fig.~\ref{fig:teaser}).
To enable the decomposition we also estimate the latent per-image illumination vectors $\vect{z}^l_j \in \mathbb{R}^{128}; j \in \{1,\ldots,q\}$~\cite{Boss2021neuralPIL}.
Furthermore, we estimate per-image camera poses and intrinsics.
Next, we provide a brief overview of prerequisites: NeRF~\cite{mildenhall2020}, InstantNGP~\cite{Mueller2022} and SAMURAI~\cite{bossSAMURAIShapeMaterial2022}.

\inlinesection{Coordinate-based MLPs and NeRF}~\cite{mildenhall2020}
uses a dense neural network to model a continuous function that takes 3D location $\vect{x} \in \mathbb{R}^{3}$ and view direction $\vect{d} \in \mathbb{R}^3$ and outputs a view-dependent output color $\vect{c} \in \mathbb{R}^{3}$ and volume density $\sigma \in \mathbb{R}$.
Mildenhall~\etal~\cite{mildenhall2020} overcome the spectral bias of the MLPs by transforming the input coordinates by a second function; A frequency encoding $\gamma$ that maps from $\mathbb{R}$ to $\mathbb{R}^{2L}$ 
~\cite{mildenhall2020,tancik2020fourfeat}:
\begin{equation}
\label{eq:fourier_enc}
\begin{split}
\gamma(\vect{x}) &= (\sin(2^0\pi \vect{x}), \cos(2^0\pi \vect{x}), \\&\ldots, \sin(2^{L-1}\pi \vect{x}), \cos(2^{L-1}\pi \vect{x}))
\end{split}
\end{equation}

\inlinesection{InstantNGP}~\cite{Mueller2022}
speed up the NeRF optimization drastically by replacing the MLP-based volume representation by a multiresolution voxel hash grid that is tailored to current GPU hardware. For a hash-size $T$, grid vertices are indexed by a spatial hash function $h(x) = \left( \underset{\scriptstyle i=1}{\overset{\scriptstyle d}{\bigoplus}} x_i \pi_i \right) \mod T$ using large unique prime numbers $\pi_i$~\cite{Mueller2022}.
At each voxel vertex a $d$-dimensional embedding is optimized.
Instead of the Fourier embedding, the 3D coordinates $\vect{x}$ are directly used to tri-linearly interpolate between neighboring vertices at each level. The results are concatenated and fed to a MLP to decode the representation. We denote the full encoding function including interpolation and concatenation as $H(\vect{x})$.

\inlinesection{Brief overview of SAMURAI.}
SAMURAI is a method for joint optimization of 3D shape, BRDF, per-image camera parameters, and illuminations for a given in-the-wild image collection.
SAMURAI~\cite{bossSAMURAIShapeMaterial2022} follows the NeRF idea outlined above but uses the Neural-PIL~\cite{Boss2021neuralPIL} method for physically-based differentiable rendering.
It takes 3D locations as input and outputs volume density and BRDF parameters. 
An additional GLO (generative latent optimization) embedding models the changes in appearances (due to different illuminations) across images.
Neural-PIL~\cite{Boss2021neuralPIL} introduced the use of per-image latent illumination embedding $\vect{z}^l_j$ and a specialized illumination pre-integration (PIL) network for fast rendering, which we refer to as `PIL rendering'. Neural-PIL optimizes a per-image embedding to model image-specific illumination.
The rendered output color $\vect{\hat{c}}$ is equivalent to NeRF's output $\vect{c}$, but due to the explicit BRDF decomposition and illumination modeling, it enables relighting and material editing.
To address the unavailability of accurate camera parameters for in-the-wild images, SAMURAI jointly optimizes camera extrinsics and per-view intrinsics from a very coarse initialization. In addition to a coarse-to-fine annealing~\cite{lin2021barf}, this is achieved 
with a multiplexed optimization scheme where multiple camera proposals per view are kept and weighted according to their performance on the loss over time. 

\vspace{-2mm}
\subsection{SHINOBI Optimization with Hash Encoding}
\label{sec:method_shinobi}
\vspace{-2mm}
We identify misaligned and inconsistent camera poses as the main limiting factor for in-the-wild reconstructions. Joint shape and camera optimization is a severely underdetermined problem. Reconstruction is typically slow and often lacks high-frequency detail in textures and shape.
Multiresolution hash grids from Instant-NGP~\cite{Mueller2022} have the potential to speed up the reconstruction while simultaneously allowing for larger ray counts to be processed and thereby improving visual quality and alignment (see Tab.~\ref{tab:3d_wild_synthesis_gt}). 
However, the naive replacement of the point encoding with Hash grids reduces the reconstruction quality and robustness of the joint camera and shape optimization. 

\noindent
Hash grids adapt to individual views faster resulting in a noisy shape in the presence of misaligned cameras. As reported previously~\cite{li2023neuralangelo,liu2023baangp,robustPoseHeo2023,zhu2023rhino} multi-resolution hash grids with the default linear interpolation 
backpropagate noisy and discontinuous gradients with respect to the input position.
Additionally, the coarse-to-fine scheme from BARF~\cite{lin2021barf} often used for camera fine-tuning cannot be directly transferred to hash grids.
Therefore, we propose an approach that makes use of a camera multiplex, adds additional geometrical constraints, and a new encoding scheme to be able to improve both reconstruction speed and quality. 
Next, we explain each of the components in detail.

\inlinesection{Architecture overview.}
A high-level overview of the~\OURS architecture is shown in \fig{overview}, which follows the skeleton of SAMURAI~\cite{bossSAMURAIShapeMaterial2022} with the `PIL renderer'~\cite{Boss2021neuralPIL}. However, we map the input coordinates $\vect{x}$ using a new hybrid encoding. 
The combined embedding is processed by a small MLP like in I-NGP~\cite{Mueller2022} to predict 
the density $\sigma$, and 
the view and appearance conditioned radiance for a given image patch. We also predict a regular direction-dependent radiance $\vect{\tilde{c}}$ to stabilize the early training stages as in~\cite{Boss2021, bossSAMURAIShapeMaterial2022}.
The  BRDF decoder operates as in SAMURAI~\cite{bossSAMURAIShapeMaterial2022}, expanding the feature representation to the BRDF (base color, metallic, roughness). Per sample, we estimate normal direction from the first order derivative of the density w.r.t.\ the input position $\frac{\partial\sigma}{\partial \vect{x}}$.
From there the volumetric rendering from NeRF~\cite{mildenhall2020} is performed and the shading for the given pixel coordinate is determined using BRDF, normals and the pre-integrated illumination estimated by the NeuralPIL network.
See supplementary material for further details on the architecture.

\inlinesection{Camera pose initialization and parameterization.}
Camera pose optimization is a highly non-convex problem and tends to quickly get stuck in local minima. 
Our initial camera poses are much noisier and feature larger distances between initial and true poses compared to many related works~\cite{wang2021nerfmm, kuang2022neroic}.
To combat this, we assume a rough initialization in the form of camera pose quadrants in line with SAMURAI~\cite{bossSAMURAIShapeMaterial2022} and NeRS~\cite{zhang2021ners}. 
We use a `lookat + direction' representation for the camera parameters, storing initial values and offsets for an eye position $\vect{p}_{eye} \in \mathbb{R}^3$, lookat direction $\Delta\vect{d}_{\phi\theta} \in \mathbb{R}^2$. and up rotation angle $d_{up} \in \mathbb{R}$ as well as the focal length $f \in \mathbb{R}$ per camera. We notice that this removes the overparameterization regarding the rotation component encoded in eye and center position of the regular `lookat' parameterization. This formulation performs best in our setting also compared to other recently proposed representations~\cite{park2023camp,zhou2019}.

\inlinesection{Hybrid positional encoding.}
We use a hash grid hybrid as coordinate encoding to improve the gradient flow \wrt the input coordinates $\vect{x}$. A Fourier-based coordinate mapping $\gamma(\vect{x})$ followed by a small MLP generates a base embedding that is concatenated with the output of the multiresolution hash grid $H(\vect{x})$ resulting in the following formulation of the neural volume $F_{\oplus}\left( (H(\vect{x}), \gamma(\vect{x}))\right)$.
On $\gamma$, we apply BARF's~\cite{lin2021barf} Fourier annealing. Similarly, we progressively add resolution levels to the hash grid encoding. Starting with only the features from a low resolution dense grid we increase the weights of the higher resolution levels gradually over time (cf.~\cite{li2023neuralangelo,liu2023baangp}.

\begin{figure}
\begin{center}

 \includegraphics[trim=0 25 0 10, clip,width=7cm] {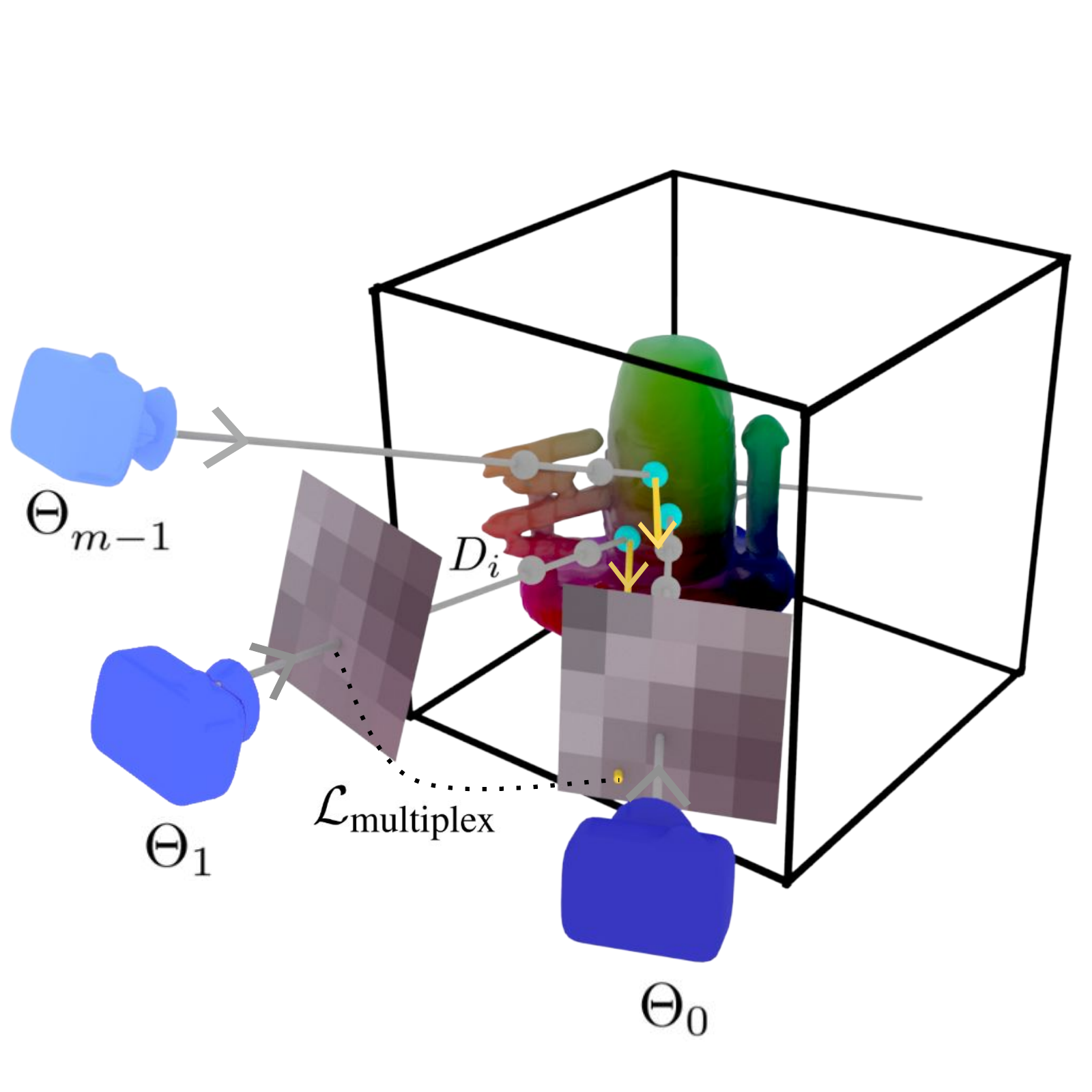}
\end{center}
\vspace{-4mm}
\titlecaption{\small Constrained camera multiplex}{\small We optimize multiple camera proposals per image and weight the contribution to the reconstruction according to a camera's performance on the loss. Between cameras of a multiplex we add a projection based regularization: Points from all members are projected into the currently best camera and then compared against a new render to enforce a consistent geometry.}
\label{fig:cam_parameterization}
\end{figure}

\inlinesection{Camera multiplexes.}
An effective way to reduce the chance of camera pose optimization to be stuck in local minima is the camera multiplex~\cite{bossSAMURAIShapeMaterial2022,goel2020cmr}.
For each image, $m$ cameras are jittered around the initial camera and simultaneously optimized. Over time the worst performing camera is repeatedly faded out until $m=1$.
This process is visualized in \fig{cam_parameterization}.
Since we render multiple proposals for a given image anyway, we see an opportunity to further constrain the optimization using projective geometry.
Specifically, we project the 2D point sets $X_i$ rendered by the $m-1$ members into the currently highest ranking camera $\Theta_{0}$ of the multiplex using the estimated depth $D_i$ from the volumetric rendering. Then we render the projected coordinates using $\Theta_{0}$ and compare the rendered color $c_i$ and alpha values $\alpha_i$ of all cameras in the multiplex to the ones originally rendered at $\Theta_{1\ldots m-1}$.
\begingroup
\setlength\abovedisplayskip{2pt}
\begin{multline}
\!\!\!\!\!\!\loss{multiplex} = \sum\limits_{i=1}^{m-1}\loss{image}(\vect{c}_i, F\hat{c}_{\mathcal{V}}\left(P_{i,0}(X_i, D_i, \Theta_i, \Theta_{0})\right)) \\
+ \loss{mask}(\vect{\alpha}_i, F\alpha_{\mathcal{V}}\left(P_{i, 0}(X_i, D_i, \Theta_i, \Theta_0)\right))
\end{multline}
\endgroup
where $P_{i, 0}$ is the perspective warp from image coordinates in camera $i$ to the reference camera. $F_{\mathcal{V}}$ is the rendering function connected to the neural field outputting color $\hat{c}$ and mask value $\alpha$, respectively.
This regularization comes roughly at the cost of adding a camera to the multiplex. Subsampling of $X_i$ can decrease the memory footprint if needed. $\loss{image}$ and $\loss{mask}$ are the optimization losses active at the time as outlined in Sec.~\ref{sec:losses_optimization}.
Naturally, this component is only active while there are multiple cameras rendered during the first part of the overall schedule. Used as an additional loss it turns out to be surprisingly effective in constraining the camera optimization and therefore increasing the robustness of the overall optimization. Essentially, we are enforcing a consistent surface to be generated and smooth the optimization landscape around an initial camera pose.

\inlinesection{View importance scaling of input images.}
Not every input might contribute to the reconstruction in the same way and individual views that are not aligned with the current 3D shape might have a negative impact on the overall optimization progress. To improve high-frequency detail in the reconstruction we reduce the impact of potentially misaligned cameras while anchoring the optimization using cameras that work well given the loss.
We keep a circular buffer of around 1000 elements with the recent per-image losses. 
Like in SAMURAI, this is used to re-weigh images in the given collection according to: $\losssuperscripted{network}{j} = {s_\text{p}}_j\,\, \losssuperscripted{network}{j}$, where
\begingroup
\small
\begin{equation}
\label{eq:post_scaling}
    {s_\text{p}}_j = \maxclip{\tanh{\left(\frac{\mu_l - ({\losssuperscripted{mask}{j}} +  {\losssuperscripted{image}{j}})}{\sigma_l}\right)} + 1}{1}\;,
\end{equation}
\endgroup
with the mean $\mu_l$ and standard deviation $\sigma_l$ of the loss buffer.
This limits the influence of badly aligned camera poses on the shape reconstruction. 
In addition, we also apply an importance weighting on $\loss{camera}$ that reduces the gradient magnitude for views that are performing well given the loss history. Specifically, at step $t$ we compute:
$\losssuperscripted{camera}{j} = {s_\text{q}}_{j,t}\,\, \losssuperscripted{camera}{j}$, with
\begingroup
\setlength\abovedisplayskip{2pt}
\small
\begin{align}
\label{eq:view_importance}
{s_\text{q}}_{j,t} =& {s_\text{q}}_{j, t-1}  \lambda_p\maxclip{\!\operatorname{tanh}{\!\left(\frac{\mu_l - ({\losssuperscripted{mask}{j}} + {\losssuperscripted{image}{j}})} {\sigma_l}\right)}\!+\!1}{1\!} \nonumber \\
+& (1-\lambda_p)  {s_\text{q}}_{j, t-1}
\end{align}
\endgroup
In practice, we set the hyperparameter $\lambda_p$ to 0.05.

\subsection{Losses and Optimization}
\label{sec:losses_optimization}
\vspace{-2mm}

\inlinesection{Multiscale patch loss}.
After a short initial phase of random ray sampling, we render randomly sampled patches of size 16x16 to 32x32. The goal is to constrain the updates and especially the alignment to be consistent on local neighborhoods. Therefore, we add a multi-scale patch loss on the rendered color $\hat{c}$ which computes a Charbonnier loss at four different resolution levels, by simple bilinear resampling. We weigh each level to compensate for the different pixel counts and enforce the low-resolution version to align first.

\begin{figure}
\begin{minipage}{\linewidth}
    \centering
     \begin{subfigure}[b]{0.32\textwidth}
         \centering
         \includegraphics[width=\textwidth]{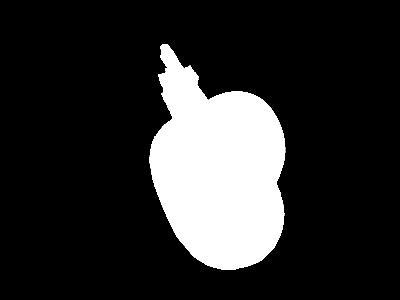}
         \caption{\footnotesize Reference silhouette}
     \end{subfigure}
     \hfill
     \begin{subfigure}[b]{0.32\textwidth}
         \centering
         \includegraphics[width=\textwidth]{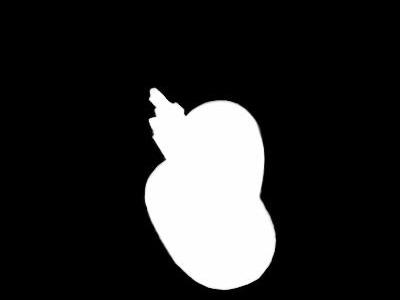}
         \caption{\footnotesize Rendered silhouette}
     \end{subfigure}
     \hfill
     \begin{subfigure}[b]{0.32\textwidth}
         \centering
         \includegraphics[width=\textwidth]{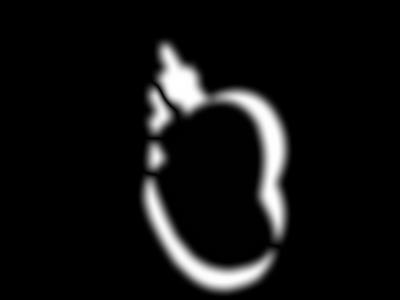}
         \caption{\footnotesize Loss map}
     \end{subfigure}
     \vspace{-2mm}
        \caption{\small \textbf{Our silhouette based alignment loss} penalizes the unaligned pixels given a reference and the rendered gray scale masks.}
        \label{fig:silhouette}
\vspace{-4mm}
\end{minipage}
\end{figure}

\inlinesection{Mask losses.}
We add a silhouette loss $\loss{Silhouette}$ whenever patch-based sampling is active. Here, we penalize the area between the two silhouettes which can be interpreted as the result of an $xor$ operation on the rendered and input mask~\cite{automatedLensch2000}. Both masks are filtered using a Gaussian blur where the radius is heuristically chosen based on the patch size. ~\fig{silhouette} visualizes how the loss helps with the alignment task. We combine this loss with a regular binary-cross-entropy loss on the mask value as well as a loss enforcing a transparent background.

\inlinesection{Regularization losses.} 
To regularize the hash grid encoding we apply a normalized weight decay as proposed in~\cite{barron2023zipnerf} to put a higher penalty on coarser grid levels compared to naive weight decay. Additionally, we apply regularization to the camera poses and normal output. Refer to the supplements for details and the hyperparameters used.

\inlinesection{Optimization.}
In total, we use three optimizers: One ADAM~\cite{kingma2014adam} optimizer for the networks, hash grid embeddings and cameras, respectively. The learning rate is decayed exponentially on all optimizers. 
In addition to the camera representation and constraints mentioned above we use ADAM with the $\beta1$ value reduced to 0.2 to smooth out the noise in the camera updates. The learning rate is tuned between \expnumber{1}{-3} to \expnumber{2}{-3} depending on scene size.
Render resolution is continuously increased over the first half of the optimization while the number of active multiplex cameras is reduced. The direct color optimization is faded to the BRDF optimization and the encoding annealing is performed over the first third of the optimization. Focal length updates and the view importance weighting are delayed until an initial shape has been formed. See the supplementary material for a detailed description and visualization of the optimization scheduling.

\vspace{2mm}

\begin{table}[ht]
\resizebox{\linewidth}{!}{ %
\Huge
\begin{tabular}{@{}
lccccccc
@{}}
\multicolumn{1}{c}{Method} & \multicolumn{2}{c}{PSNR\textuparrow} & \multicolumn{2}{c}{SSIM\textuparrow}  & \multicolumn{2}{c}{LPIPS\textdownarrow} & \multicolumn{1}{c}{Runtime}
\\
\cmidrule(lr){2-3}
\cmidrule(lr){4-5}
\cmidrule(lr){6-7}
 & 
 $S_C$ & $\sim S_C$ & $S_C$ & $\sim S_C$ & $S_C$ & $\sim S_C$ &
\\
\midrule
NeROIC~\cite{kuang2022neroic}  & 
22.75 & 21.31 & 0.91 & 0.90 & 0.0984 & \secondbest{0.0845} & 18 hours (4 GPUs)
\\
NeRS~\cite{zhang2021ners}  & 
17.92 & 18.02 & \secondbest{0.92} & \secondbest{0.93} & 0.114 & 0.1098 & 3 hours (1 GPU)
\\
SAMURAI~\cite{bossSAMURAIShapeMaterial2022}  &
 \secondbest{25.34} & \secondbest{24.61} & \secondbest{0.92} & 0.91 & \secondbest{0.0958} & 0.1054 & 12 hours (1 GPU)
\\
\OURS  &
 \best{27.69} & \best{27.79} & \best{0.94} & \best{0.94} & \best{0.0607} & \best{0.0578} & 4 hours (1 GPU)
\\
\end{tabular}
} %
\begingroup %
\titlecaptionof{table}{\small Metrics for view synthesis on NAVI}{\small View synthesis metrics are computed over two subsets from all wild-sets depending on the success of COLMAP ($S_C$ / $\sim S_C$). Rendering quality is evaluated on a holdout set of test views. We initialize with the GT poses provided by NAVI~\cite{jampani2023navi}. }
\label{tab:3d_wild_synthesis_gt}
\endgroup
\end{table}

\inlinesection{Implementation.}
We implement the multi-resolution hash grid encoding as a custom CUDA extension for Tensorflow~\cite{tensorflow2015}. 
The implementation roughly follows the official CUDA implementation~\cite{Mueller2022}.  
We enable first- and second-order gradients for the encoding to allow for computing analytical surface normals.
The remaining components are implemented in Tensorflow.

\vspace{2mm}

\begin{table*}[ht]
\scriptsize
\begin{tabular}{@{}
llcccccc
S[table-format=1.2(3)]
S[table-format=1.2(3)]
S[table-format=3.2(4)]
S[table-format=3.2(4)]
@{}}
\multicolumn{1}{c}{Method} & \multicolumn{1}{c}{Pose Init} & \multicolumn{2}{c}{PSNR\textuparrow} & \multicolumn{2}{c}{SSIM\textuparrow}  & \multicolumn{2}{c}{LPIPS\textdownarrow}
& \multicolumn{2}{c}{Translation\textdownarrow} 
& \multicolumn{2}{c}{Rotation \si{\degree} \textdownarrow}
\\
\cmidrule(lr){3-4}
\cmidrule(lr){5-6}
\cmidrule(lr){7-8}
\cmidrule(lr){9-10}
\cmidrule(lr){11-12}
 & & 
 $S_C$ & $\sim S_C$ & $S_C$ & $\sim S_C$ & $S_C$ & $\sim S_C$ &
 {$S_C$} & {$\sim S_C$} & {$S_C$} & {$\sim S_C$}
\\
\midrule
GNeRF~\cite{meng2021gnerf} & Random &
8.30 & 6.25 & 0.64 & 0.63 & 0.52 & 0.57 & 1.02\pm 0.16 & 1.04\pm 0.09 & 93.15\pm 26.54 & 80.22\pm 27.64
\\ \midrule
NeROIC~\cite{kuang2022neroic} & COLMAP & 
19.77 & - & 0.88 & - & 0.150 & - & 0.09\pm 0.12 & {-} & 42.11\pm 17.19 & {-} 
\\
NeRS~\cite{zhang2021ners} & Directions & 
18.67 & 18.66 & \best{0.92} & \best{0.93} & 0.108 & 0.107 & 0.49\pm 0.21 & 0.52\pm 0.19 & 122.41\pm 10.61 & 123.63\pm 8.8 
\\
SAMURAI~\cite{bossSAMURAIShapeMaterial2022} & Directions &
 \best{25.34} & \secondbest{24.61} & \best{0.92} & 0.91 & \secondbest{0.096} & \secondbest{0.105} & \best{}0.24\pm 0.17 & \secondbest{}0.35\pm 0.24 & \secondbest{}26.16\pm 22.72 & \secondbest{}36.59\pm 29.98
\\ \midrule
\OURS & Directions &
 \secondbest{25.15} & \best{24.77} & \best{0.92} & \secondbest{0.92} & \best{0.090} & \best{0.095} & \secondbest{}0.25\pm 0.085 & \best{}0.28\pm 0.09 & \best{}22.84\pm 16.19 & \best{}33.00\pm 19.97
\end{tabular}
\begingroup %
\titlecaptionof{table}{\small Metrics for 3D shape and pose on NAVI}{\small View synthesis and pose metrics over two subsets from all wild-sets depending on the success of COLMAP ($S_C$ / $\sim S_C$). Rendering quality is evaluated on a holdout set of test views that are aligned as part of the optimization without contributing to the shape recovery. We include GNeRF as a separate baseline although this method is not designed for multi-illumination data. We report metrics with the methods' default camera initialization and evaluate against the annotation provided in NAVI~\cite{jampani2023navi}.}
\label{tab:3d_wild_synthesis}
\endgroup
\end{table*}

\vspace{2mm}

\vspace{2mm}
  
\!\!\!\!\!\!\!\resizebox{0.98\linewidth}{!}{ %
\Huge
\begin{tabular}{@{}lcccc@{}}
Method & PSNR\textuparrow & SSIM\textuparrow & Transl.\textdownarrow &
Rot. \textdegree \textdownarrow
\\
\midrule
w/o Multiplex Consistency Loss& 25.80 & \best{0.93} & \best{0.29} & 23.12\\
w/o Per View Importance & 22.43 & 0.90 & 0.36 & 35.10\\
w/o Coarse-to-fine (annealing) & 21.47 & 0.90 & 0.37 & 30.44\\
w/o Hybrid Encoding & 25.31 & \best{0.93} & 0.30 & 23.33 \\
w/o Patch-based Training & 20.60 & 0.89 & 0.45 & 41.30\\
\textbf{Full} & \best{25.87} & \best{0.93} & 0.30 & \best{22.90}
\end{tabular}
} %
\titlecaptionof{table}{Ablation study}{Ablating components of our framework results in worse view synthesis and relighting results (averaged over "Keywest" and "School Bus" scenes from NAVI) demonstrating their importance.
}
\label{tab:ablations}

\vspace{6mm}

\section{Experiments}
\label{sec:results}
\vspace{-2mm}
\inlinesection{Dataset} 
For evaluations, we use the in-the-wild collections from the NAVI dataset~\cite{jampani2023navi} which feature objects captured in diverse environments using multiple mobile devices. High-quality annotated camera poses allow us to ablate and perform quantitative evaluation of our pose estimation.

\inlinesection{Baselines.} The closest prior work that can tackle 
our task outline in Sec.~\ref{sec:method} is SAMURAI~\cite{bossSAMURAIShapeMaterial2022} on which our method is based. We compare against SAMURAI as a baseline and also conduct experiments using NeROIC~\cite{kuang2022neroic}, GNeRF~\cite{meng2021gnerf}, and a modified version of NeRS~\cite{zhang2021ners} (details in the supplement). For experiments on joint shape and pose estimation, we use the same quadrant-based pose initialization for NeRS, SAMURAI and SHINOBI (ours);
and we use the the methods' default pose initializations for NeROIC (COLMAP) and GNeRF (Random).

\inlinesection{Evaluation.} 
We use two strategies for evaluation. First, the standard novel view synthesis metrics using the learned volumes that measure PSNR, SSIM, and LPIPS~\cite{zhang2018perceptual} scores on held-out test images.
Second, to evaluate camera poses w.r.t GT poses, we use Procrustes analysis~\cite{gower2004procrustes} to align the cameras and then compute the mean absolute rotation and translation differences in camera pose estimations for all available views.
For evaluation purposes, we optimize the cameras and illuminations on the test images but do not allow the test images to affect the other network parts or hash grid embedding.
For a fair comparison, we use the ground truth masks as input to all methods although our method also includes functionality to automatically generate segmentation masks. We run experiments on a single Nvidia A100 or V100 GPU per scene.
\begin{table*}[t]
    \input{figs/decomposition_qualitative}
\end{table*} 

\inlinesection{Results.} 
Tab.~\ref{tab:3d_wild_synthesis_gt} shows the performance of different methods for in-the-wild reconstruction when using GT poses from NAVI. 
Following NAVI~\cite{jampani2023navi}, we divide the scenes into two subsets based on whether the COLMAP works ($S_C$) or not ($\sim S_C$) as some techniques like NeROIC need COLMAP poses to work on unposed image collections.
Using the provided annotated poses~\OURS clearly performs best on the view synthesis task (Tab.~\ref{tab:3d_wild_synthesis_gt}). This shows the advantage of our hybrid encoding scheme and the patch-based losses over previous methods for in-the-wild scenes. Optimization runtimes of different techniques show that we are 3 times faster than the next-best SAMURAI approach.

\begin{figure}
\begin{minipage}{\linewidth}
\includegraphics[width=1.1\textwidth]{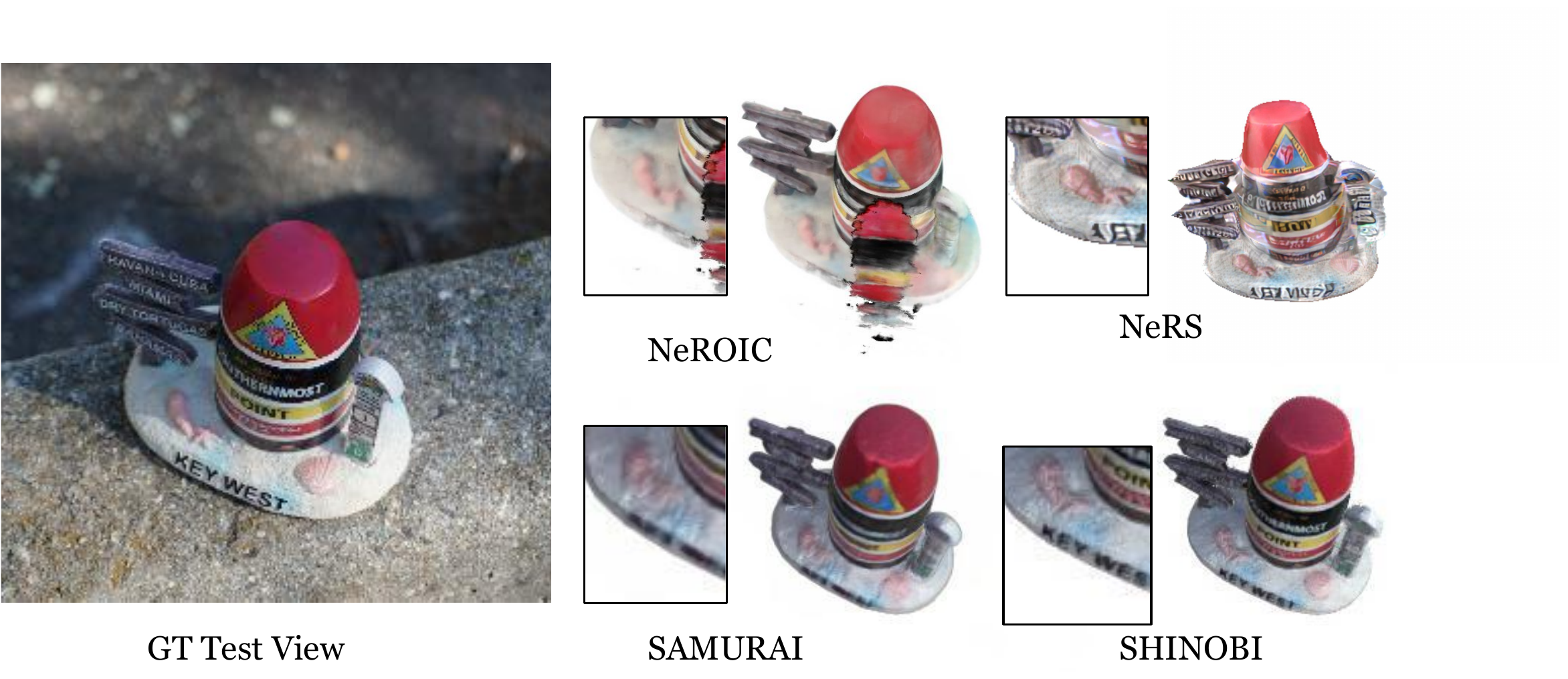} 
\caption{\small \textbf{Novel view synthesis compared to existing methods.} Compared to other methods on an example view from the NAVI~\cite{jampani2023navi} in-the-wild test set, SHINOBI preserves fine detail and recreates the lighting realistically.}
\label{fig:comparison_view_synthesis}
\end{minipage}
\end{figure}

\begin{figure}
\begin{minipage}{\linewidth}
    \centering
     \begin{subfigure}[b]{0.32\textwidth}
         \centering
         \includegraphics[width=\textwidth]{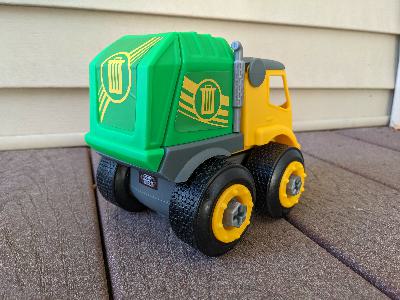}
     \end{subfigure}
     \hfill
     \begin{subfigure}[b]{0.32\textwidth}
         \centering
         \includegraphics[width=\textwidth]{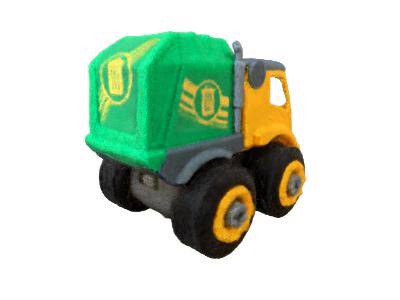}
     \end{subfigure}
     \hfill
     \begin{subfigure}[b]{0.32\textwidth}
         \centering
         \includegraphics[width=\textwidth]{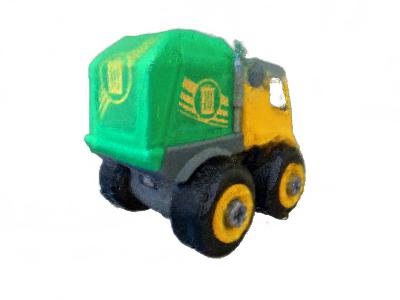}
     \end{subfigure}
     \\
     \vspace{-1mm}
     \begin{subfigure}[b]{0.32\textwidth}
         \centering
         \includegraphics[trim=50 50 60 20,clip,width=\textwidth]{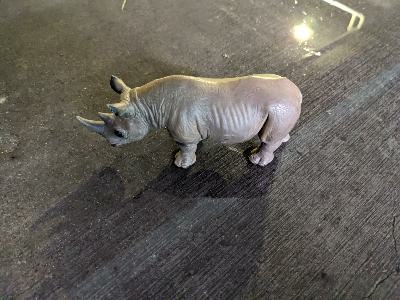}
     \end{subfigure}
     \hfill
     \begin{subfigure}[b]{0.32\textwidth}
         \centering
         \includegraphics[trim=50 50 60 20,clip,width=\textwidth]{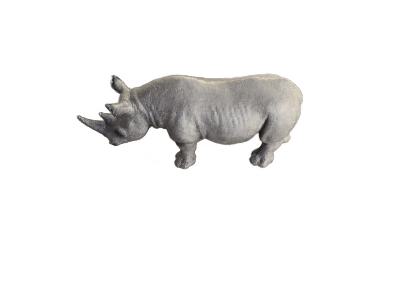}
     \end{subfigure}
     \hfill
     \begin{subfigure}[b]{0.32\textwidth}
         \centering
         \includegraphics[trim=50 50 60 20,clip,width=\textwidth]{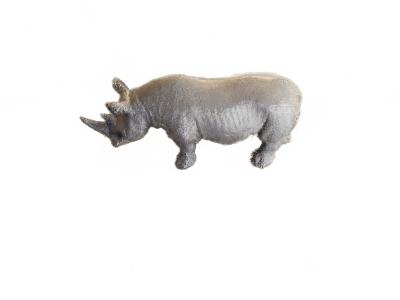}
     \end{subfigure}
     \\
     \vspace{-1mm}
     \begin{subfigure}[b]{0.32\textwidth}
         \centering
         \includegraphics[trim=10 10 10 10,clip,width=\textwidth]{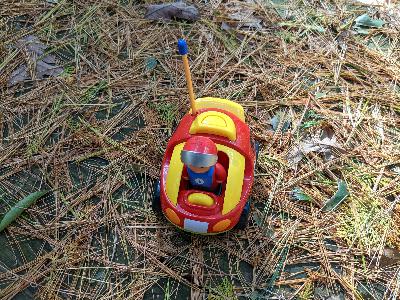}
         \caption{\footnotesize GT Novel View}
     \end{subfigure}
     \hfill
     \begin{subfigure}[b]{0.32\textwidth}
         \centering
         \includegraphics[trim=10 10 10 10,clip,width=\textwidth]{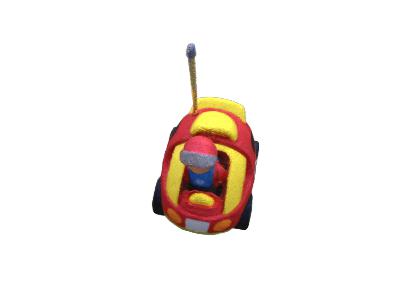}
         \caption{\footnotesize GT pose init.}
     \end{subfigure}
     \hfill
     \begin{subfigure}[b]{0.32\textwidth}
         \centering
         \includegraphics[trim=10 10 10 10,clip,width=\textwidth]{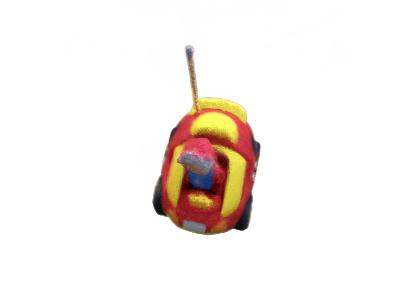}
         \caption{\footnotesize Direction pose init.}
     \end{subfigure}
     \vspace{-2mm}
        \caption{\small \textbf{View synthesis on NAVI.} Renderings from~\OURS using models initialized with camera pose quadrants only or the GT provided by NAVI~\cite{jampani2023navi} compared to the input image.}
        \label{fig:navi_initializations}
\vspace{-4mm}
\end{minipage}
\end{figure}

~~Tab.~\ref{tab:3d_wild_synthesis} shows results of joint shape and pose optimization from in-the-wild image collections when the GT camera poses are not given as input. SHINOBI outperforms both NeROIC and NeRS by a healthy margin while being on-par with SAMURAI.
While PSNR of SHINOBI is similar to SAMURAI, our method is able to reconstruct scenes consistently with lower translation and rotation pose errors (with also lower standard deviation in pose metrics).
This results in SHINOBI obtaining better LPIPS perceptual metrics compared to SAMURAI.
The on-par mean PSNR compared to SAMURAI mostly stems from individual test cameras not being aligned properly. This also happens for other methods but seems to be emphasized by the faster optimization scheduling in SHINOBI. 
NeROIC can also achieve good results if camera poses are close to the ground truth but fails for many scenes where a COLMAP-based initialization is not possible. NeRS also succeeds in reconstructing all scenes. However, it achieves lower-quality camera alignments.
Fig.~\ref{fig:comparison_view_synthesis} visually compares view synthesis results from different methods, which visually confirms that SHINOBI can produce sharper results that are more faithful to the input images.
Further results on the NAVI dataset are shown in \fig{navi_initializations}, where we show novel views predicted by~\OURS initialized with either GT poses or rough quadrants. Visual results clearly show that \OURS can recover the pose and provide a consistent illumination w.r.t the ground-truth target views in both settings.

\inlinesection{Decomposition results.}
Fig~\ref{fig:decomposition_comparison} compares the BRDF and illumination decomposition of~\OURS to SAMURAI where the same output modalities are available. Visual results show significantly more high-frequency detail and plausible material parameters with~\OURS compared to SAMURAI.

\inlinesection{Ablation study.} 
We ablate different aspects of~\OURS in terms of reconstruction metrics using 
the ``Keywest" and ``School Bus", two in-the-wild sets from NAVI~\cite{jampani2023navi} of medium complexity. %
Metrics in Tab.~\ref{tab:ablations} show that the resolution annealing coarse-to-fine scheme and the patch-based losses contribute most significantly to the final quality. The latter improves local details and registration accuracy compared to a simple pixel-wise loss. The view importance weighting is another important factor for improved sharpness. It helps to stabilize the optimization after the initial resolution annealing schedule has ended.
While the hybrid encoding and camera multiplex consistency do not seem to have a large impact quantitatively, they play a critical role in stabilizing the optimization over different scene types and scales. Without them, the optimization might take longer or diverge depending on the initialization. Visual examples of the specific ablations are compared in the supplementary material.

\inlinesection{Applications.}
In addition to novel view synthesis using the NeRF~\cite{mildenhall2020} representation, the parametric material model allows for controlled editing of the object's appearance. Also the illumination can be adjusted, e.g.\ for realistic composites. A mesh extraction allows further editing and integration in the standard graphics pipeline including real-time rendering. %
~\OURS can help in obtaining relightable 3D assets for e-commerce applications as well as 3D AR and VR for entertainment and education. Refer to the supplementary material for sample visual results on relighting, material editing etc.

\begin{figure}
\begin{minipage}{\linewidth}
    \centering
     \begin{subfigure}[b]{0.45\textwidth}
         \centering
         \includegraphics[width=\textwidth]{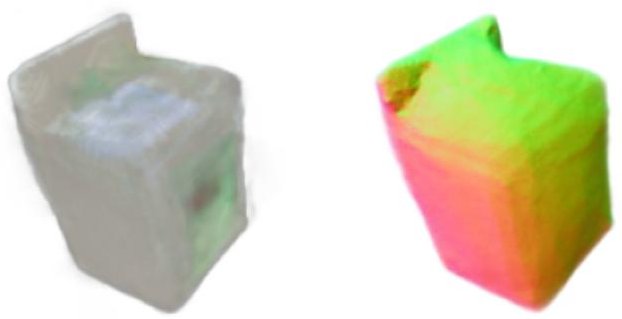}
         \caption{\footnotesize Kitchen sink (rgb, normals)}
     \end{subfigure}
     \hfill
     \begin{subfigure}[b]{0.53\textwidth}
         \centering
         \includegraphics[width=0.45\textwidth]{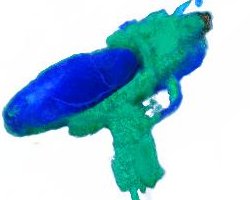}\quad
         \includegraphics[width=0.45\textwidth]{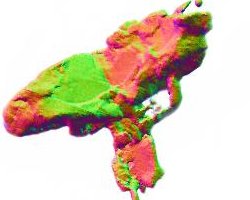}
         \caption{\footnotesize Water gun (rgb, normals)}
     \end{subfigure}
         
     \vspace{-2mm}
        \caption{\small \textbf{Failure cases.} Unconstrained image collections featuring highly symmetric objects or homogenous surfaces still pose a challenge and potentially require additional assistance.}
        \label{fig:failures}
\vspace{-4mm}
\end{minipage}
\end{figure}

\inlinesection{Limitations.}
Joint pose and shape reconstruction is an inherently ill-posed problem. While~\OURS improves over previous work, especially symmetric objects and highly specular materials can lead to failure cases as shown in 
Fig.~\ref{fig:failures}. The coarse-to-fine scheme is not able to resolve the disambiguities and the camera poses are stuck in a local minimum. All existing methods show these limitations to some extent. In some regions, high-frequency detail is still not reconstructed properly due to misaligned views and the band limited capabilities of the illumination representation~\cite{Boss2021neuralPIL}. Furthermore, our BRDF and illumination decomposition is not capable of modeling shadowing and inter-reflections. As we are mainly concerned with single-object decomposition, 
these are not crucial. Extending this method to more complex light transport modeling forms an important future work. %

\vspace{-2mm}
\section{Conclusion}
\label{sec:conclusion}
\vspace{-2mm}
We present SHINOBI, a framework for shape, pose, and illumination estimation of objects from unposed in-the-wild image collections. Using a hybrid hash grid encoding scheme we enable easier camera pose optimization using a multiresolution hash grid. Additionally, our choice of camera parameterization along with per-view importance weighting and patch-based alignment loss allows for a better image-to-3D alignment resulting in better reconstruction with high-frequency details. Although SHINOBI is able to recover the geometry of objects from any category, its performance is limited on thin/transparent structures and fails to recover high-frequency details under extreme illumination changes, which we leave as exploration for future work.

\ifreview
\else
\section*{Acknowledgements}
\vspace{-2mm}
This work has been partially funded by the Deutsche Forschungsgemeinschaft (DFG, German Research Foundation) under Germany’s Excellence Strategy – EXC number 2064/1 – Project number 390727645 and SFB 1233, TP 02  -  Project number 276693517.
\fi

{\small
\bibliographystyle{ieeenat_fullname}
\bibliography{11_references}
}

\ifarxiv \clearpage \appendix \section*{Overview}
\ifreview
In the supplement to SHINOBI, a method for 3D joint reconstruction of shape, illumination and materials from in-the-wild image sequences, we first present additional details on the method's architecture (Sec.\ref{sec:sup_method_details_architecture}) and the optimization (Sec.~\ref{sec:sup_method_details_optimization}). In Sec.~\ref{sec:sup_experiments} we introduce additional qualitative results from object reconstructions of the NAVI dataset~\cite{jampani2023navi} and add visual examples to our ablation study. Finally, applications of our reconstructed data are shown in Sec.~\ref{sec:sup_exp_applications}.
Please also consider watching the \textbf{supplemental video} for an overview of this work and further visual results.
\fi
\ifarxiv
In the supplement to SHINOBI, a method for 3D joint reconstruction of shape, illumination and materials from in-the-wild image sequences, we first present additional details on the method's architecture (Sec.\ref{sec:sup_method_details_architecture}) and the optimization (Sec.~\ref{sec:sup_method_details_optimization}). In Sec.~\ref{sec:sup_experiments} we introduce additional qualitative results from object reconstructions of the NAVI dataset~\cite{jampani2023navi} and add visual examples to our ablation study. Finally, applications of our reconstructed data are shown in Sec.~\ref{sec:sup_exp_applications}.
Please also visit our \textbf{project page} for an overview of this work and further visual results video.
\fi
\ifcamera
In the supplement to SHINOBI, a method for 3D joint reconstruction of shape, illumination and materials from in-the-wild image sequences, we first present additional details on the method's architecture (Sec.\ref{sec:sup_method_details_architecture}) and the optimization (Sec.~\ref{sec:sup_method_details_optimization}). In Sec.~\ref{sec:sup_experiments} we introduce additional qualitative results from object reconstructions of the NAVI dataset~\cite{jampani2023navi} and add visual examples to our ablation study. Finally, applications of our reconstructed data are shown in Sec.~\ref{sec:sup_exp_applications}.
Please also visit our \textbf{project page} for an overview of this work and further visual results video.
\fi

\begin{figure*}[ht]
\begin{minipage}{\linewidth}
\centering
     \begin{subfigure}[b]{0.32\textwidth}
         \centering
         \includegraphics[width=\textwidth]{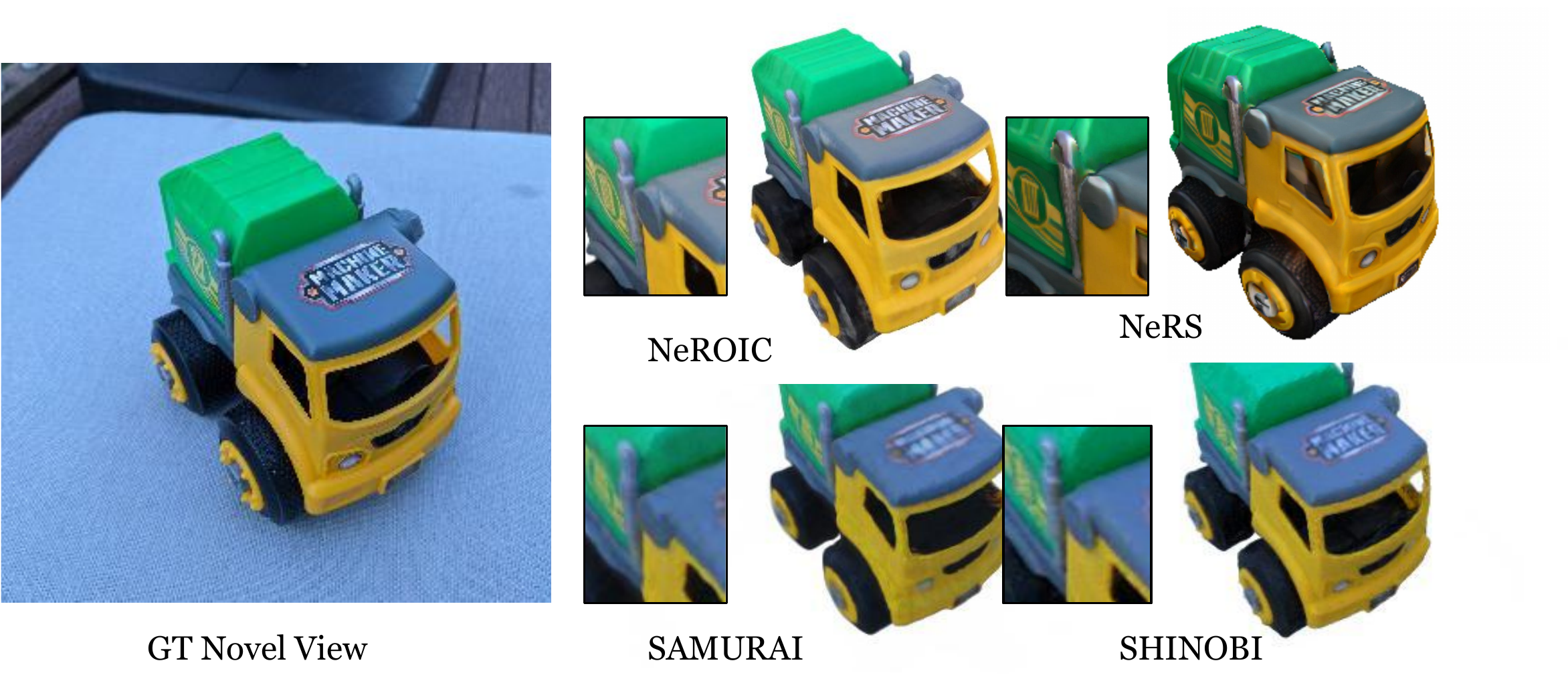}
     \end{subfigure}
     \hfill
     \begin{subfigure}[b]{0.32\textwidth}
         \centering
         \includegraphics[width=\textwidth]{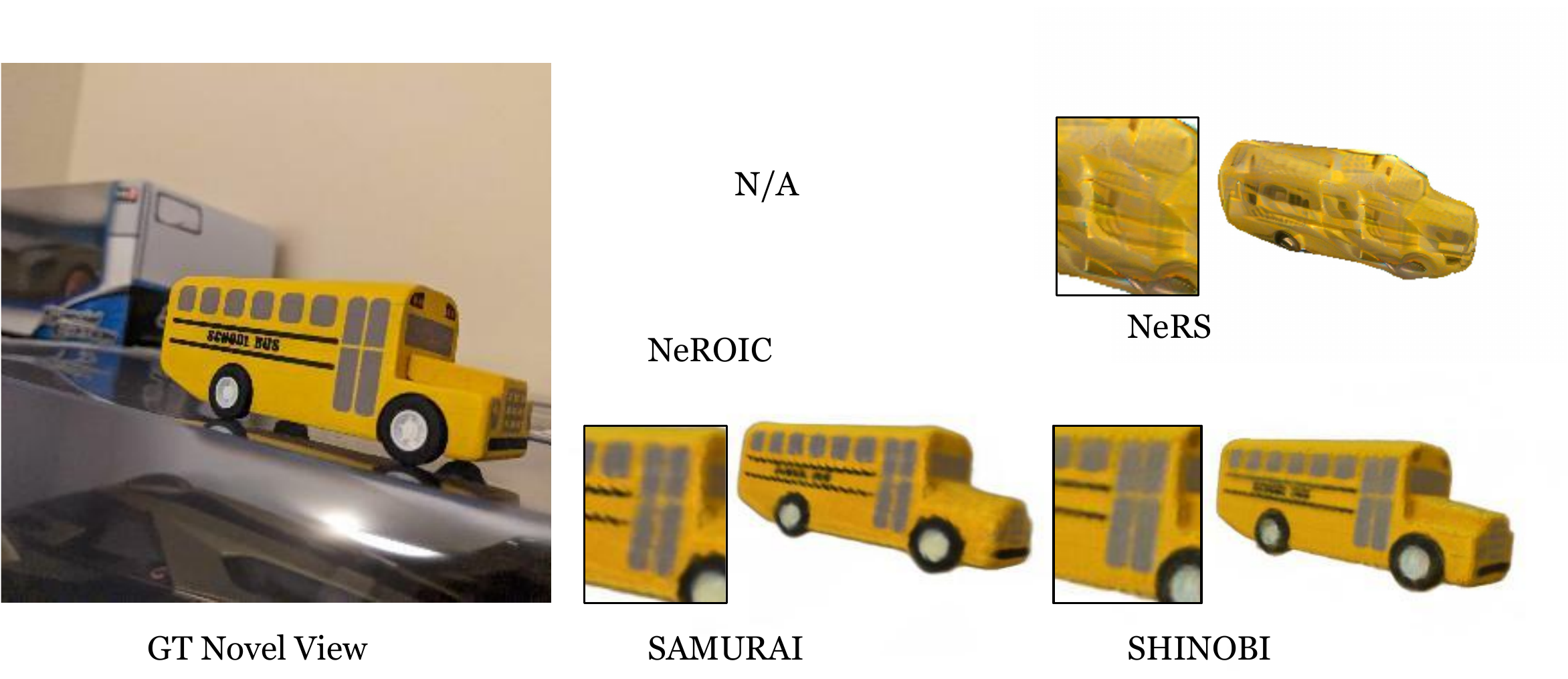}
     \end{subfigure}
     \hfill
     \begin{subfigure}[b]{0.32\textwidth}
         \centering
         \includegraphics[width=\textwidth]{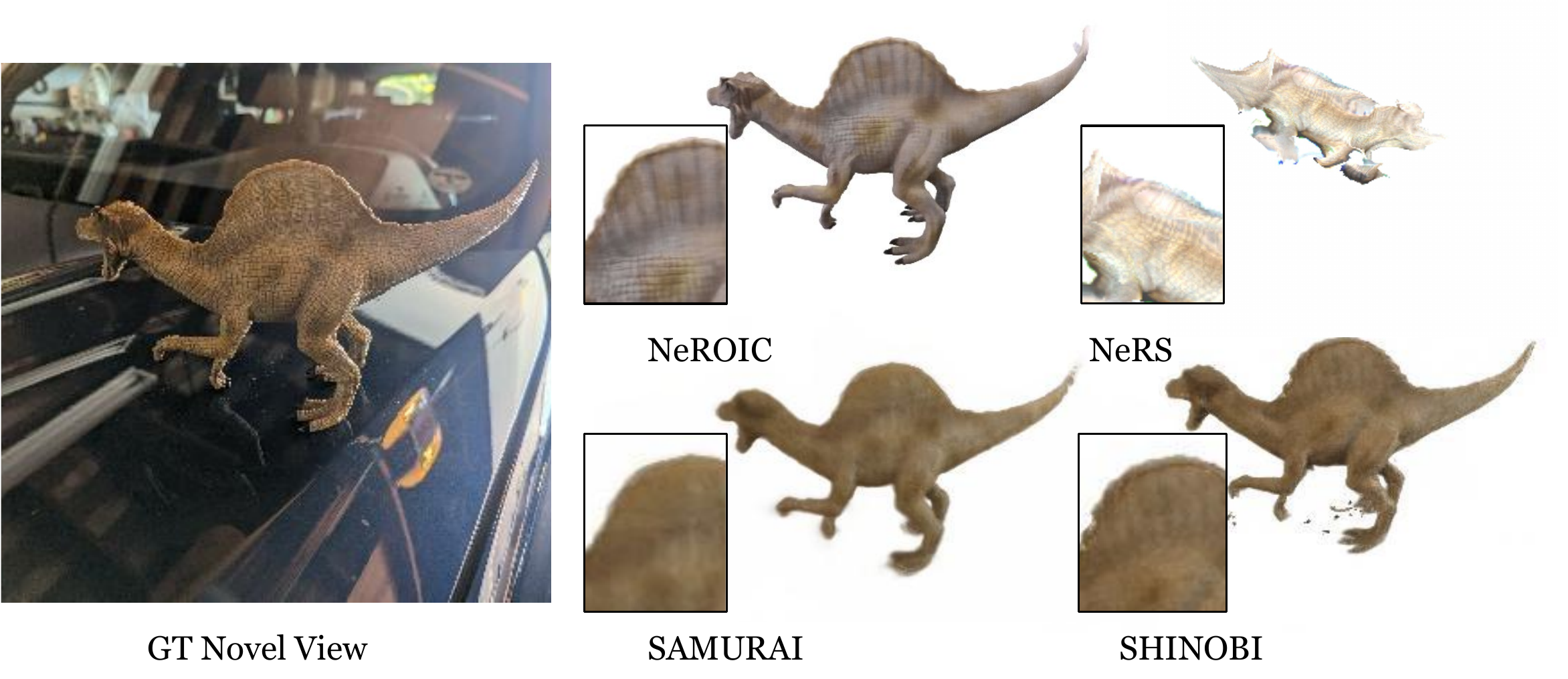}
     \end{subfigure}
\caption{\small \textbf{Novel view synthesis compared to existing methods.} Additional example objects from NAVI~\cite{jampani2023navi} in-the-wild image collections. SHINOBI robustly reconstructs even when initialized with exteremely coarse poses while e.g. NeROIC~\cite{kuang2022neroic} does not succeed on some scenes.}
\label{fig:sup_comparison_view_synthesis}
\end{minipage}
\end{figure*}

\section{Additional Method Details}
\label{sec:sup_method_details}

\subsection{NeRF Raymarching}
As introduced in Sec.~\ref{sec:method} the neural networks in NeRF~\cite{mildenhall2020} output a vector for view-dependent output color $\vect{c} \in \mathbb{R}^{3}$ and volume density $\sigma \in \mathbb{R}$ given a 3D location $\vect{x} \in \mathbb{R}^{3}$ and view direction $\vect{d} \in \mathbb{R}^3$.
A camera ray $r(t) = \vect{o} + t \vect{d}$ is cast into the volume, with ray origin $\vect{o} \in \mathbb{R}^3$ and view direction $\vect{d}$. 
The final color is then approximated via numerical quadrature of the integral: $\vect{\hat{c}}(\vect{r})=\int_{t_n}^{t_f} T(t)\sigma(t)\vect{c}(t)\, dt$ with $T(t) = \exp (-\int_{t_n}^{t} \sigma(t)\, dt )$, using the near and far bounds of the ray $t_n$ and $t_f$ respectively~\cite{mildenhall2020}.
Originally, the first MLP learns a coarse representation by sampling the volume in a fixed uniform sampling pattern along each ray. The second MLP is then evaluated sampled according to the coarse density distribution, placing more samples in high-density areas.
In SHINOBI we only use one sampling stage with uniformly samples along rays.
Using a raymarching scheme that skips empty space based iteratively updated occupancy data could bring additional performance gain during optimization.

\subsection{Architecture}
\label{sec:sup_method_details_architecture}
\inlinesection{Hybrid hash encoding configuration.}
The hybrid encoding features two branches.
For the base encoding we use 10 random offset annealed Fourier frequencies for the positional encoding followed by a small MLP featuring a single hidden layer with $64$ dimensions and silu activation~\cite{Elfwing2017silu}. The output equals the input dimension ($3$), again as it is done by Zhu~\etal~\cite{zhu2023rhino}.
We apply BARF's~\cite{lin2021barf} Fourier annealing and add random frequencies as offsets to the logarithmically spaced frequencies~\cite{tancik2020fourfeat, bossSAMURAIShapeMaterial2022} to prevent artifacts from axis-aligned frequencies.
The multiresolution hash grid is configured with $16$ levels with a base resolution of $8$ and a maximum target resolution of $2048$. The embedding dimensions are $2$ or $4$. The experiments reported in Sec. 4 of our paper are generated using $2$ dimensions. A slightly better decomposition quality can be achieved by increasing the dimensionality at the cost of increased memory consumption and runtime.
Hence, the final feature dimension after encoding and concatenation is $35$ or $67$. See Sec.~\ref{sec:hashgrid_annealing} for an explanation of the annealing strategy applied to the hash grids.

\inlinesection{Networks.}
The main network taking in the encoded features consists of 3 ReLU~\cite{hahnloser2000relu} activated layers with 64 channels. An additional linear layer generates the output for the $\sigma$ density parameter from the 64 channel activation. Softplus $\mathrm{softplus}(x) = \ln(1 + {\rm e}^x)$~\cite{Dugas2000_softplus} is applied to the raw $\sigma$.
The directions are encoded using 4 non-annealed regular Fourier components as in Mildenhall~\etal~\cite{mildenhall2020} and then, concatenated with the main network output, fed to a secondary MLP to predict the view direction-dependent radiance $\vect{\tilde{c}}$ used in the beginning of the optimization. The secondary conditional network has a hidden dimension of 32 in our case.
For the BRDF prediction a single linear layer compresses the main network output to 16 channels. From there the BRDF decoder is applied which consists of another two layers with 64 channels and ReLU activation each. Each BRDF output; basecolor, metallic and roughness has its own output layer followed by a sigmoid activation~\cite{Boss2021neuralPIL}. An additional diffuse embedding is added as conditioning to the basecolor branch before output.
The illumination network decoding the per view latent vector is conditioned by the same configuration of mapping layers as outlined in Neural-PIL~\cite{Boss2021neuralPIL}. 

\inlinesection{Multiresolution hash grid level annealing.}
\label{sec:hashgrid_annealing}
Inspired by BARF~\cite{lin2021barf} and Nerfies~\cite{park2020nerfies} we apply a coarse-to-fine annealing to the hash grid encoding by weighting the different grid levels. Starting with only the features from the low resolution dense grid and all other features set to zero we increase the weights of the higher resolution levels gradually over time (cf.~\cite{li2023neuralangelo,liu2023baangp}). Similar to the implementation by Lin~\etal we formulate it as a truncated Hann window:
\begin{align}
    \Gamma_k(\vect{x}; \alpha) &= w_k(\alpha) \left[ \sin(2^k \vect{x}), \cos(2^k \vect{x}) \right] \\
    w_k(\alpha) &= \frac{1 - \cos \left(\pi\, \text{clamp}(\alpha - k, 0, 1)  \right)}{2}
\end{align}
where $\alpha \in [0, L]$ with $L$ being the number of resolution levels of the hash grid encoding.

We also tested the idea of BAA-NGP~\cite{liu2023baangp} replicating embeddings from low-resolution levels but observed reduced performance in our optimization setting.
Similarly, we had no success with adding a straight-through operator to the interpolation on the hash grid as proposed in~\cite{robustPoseHeo2023}.

\subsection{Camera Parameterization.}
We label initial poses based on 3 simple binary questions: Left \vs Right, Above \vs Below, and Front \vs Back. This only takes about 4-5 minutes for a typical 80 image collection. Alternatively, our framework allows to extend the initialization to a camera multiplex spanning more than one quadrant. This can enable fully random initialization for front-facing scenes and image sets featuring rotating cameras with a fixed object distance as shown by Levy~\etal~\cite{levyMELONNeRFUnposed2023}. As these constrained settings are uncommon for in-the-wild collections we discard it here.
We use a perspective pinhole camera model and an initial field of view of 53.13 degrees. 
We optimize offsets to the original camera parameters of our `lookat + direction' parameterization as outlined in the main paper. Here, we encode the trainable lookat parameter $\Delta\vect{d}$ directly as two direction components, $\phi$, $\theta$, which are used to offset the viewing direction $\vect{d}$ to obtain the updated $\hat{\vect{d}}$ as follows:
\begin{align}
   \vect{d} &= (\vect{p}_{\text{eye}} + \Delta\vect{p}_{\text{eye}}) - \vect{p}_{\text{center}} \\
   \theta &= \arcsin(\vect{d}_y) + \Delta\vect{d}_{\theta}\\
   \phi &= \arctantwo(\vect{d}_x, \vect{d}_z) + \Delta\vect{d}_{\phi}\\
   \hat{\vect{d}} &= \langle \cos{\phi} \sin{\theta}, \sin{\phi}, \cos{\phi} \cos{\theta} \rangle
\end{align}
We limit $\Delta\vect{d}$ to the range $\left[ -0.5 \pi, 0.5 \pi\right]$.

We also tried other camera parameterizations like the popular 6D rotation representation by Zhou~\etal~\cite{zhou2019} or FocalPose~\cite{ponimatkin2022focalpose} that has recently been applied to NeRF with camera fine-tuning~\cite{park2023camp}. Interestingly, our lookat + direction parameterization performs the best in our setting as it seems to work well with the regularizations on camera poses.

\subsection{Regularization and Losses.}
\inlinesection{Multiresolution hash grid regularization.}
To regularize the hash grid encoding we use the following normalized weight decay as proposed by Barron~\etal~\cite{barron2023zipnerf}: $\loss{Grid} = \sum_l \text{mean}(V_l)$ with $V_l$ referring to the grid embeddings at resolution level $l$. Computing the sum of the mean per-level puts a higher penalty on coarser grid levels compared to naive weight decay over all parameters at once. We find a weighting of 0.02 to 0.05 work well in our setting and settle for 0.02 as the final value.
We apply gradient scaling to the gradients for the network by the norm of $0.1$.
Furthermore, gradient norm clipping with a clip value of $2.5$ is applied to the camera gradients before the parameter update.

\inlinesection{Surface normals regularization.}
We use the normal direction loss $\loss{ndir}$ from~\cite{Verbin2022} to constrain the normals to face the camera until the ray reaches the surface. This helps in providing sharper surfaces without floater artifacts. Additionally, we observe that the explicit rendering step helps to constrain the surface normals as noise is reduced compared to optimization using only the predicted radiance.

\inlinesection{Camera regularization.}
The camera regularization losses from SAMURAI are kept, particularly one to force the lookat-direction to point towards the origin ($\loss{Lookat}$) and one to prevent the cameras from moving too far away from the bounding volume ($\loss{Bounds}$)~\cite{bossSAMURAIShapeMaterial2022}. An additional term on the magnitude of the camera offset parameters helps to keep cameras from moving too far too fast with respect to the initial position due to strong updates in the beginning of the optimization.

\inlinesection{BRDF losses.}
Joint estimation of BRDF and illumination is a delicate endeavor. For example, the illumination can easily fall into a local minimum. The object is then tinted in a bluish color, and the illumination is an orange color to express a more neutral color tone, for example. As our image collections have multiple illuminations, we can force the base color $\vect{b}_c$ to replicate the pixel color from the input images. This way, a mean color over the dataset is learned and it becomes less likely to be trapped in local minima. We evaluate the Mean Squared Error (MSE) for this: $\loss{Init} = \loss{MSE}(\vect{C^s}, \vect{b}_c)$. Additionally, we add a smoothness loss $\loss{Smooth}$ for the normal, roughness, and metallic parameters similar to the one used in UNISURF~\cite{Oechsle2021} to further regularize BRDF estimation~\cite{bossSAMURAIShapeMaterial2022}.

\inlinesection{Image reconstruction loss} is a Charbonnier loss: $\loss{Image}(g, p) = \sqrt{(g - p)^2 + 0.001^2}$ between the input color from $C$ for pixel $s$ and the corresponding predicted color of the networks $\vect{\tilde{c}}$. We also calculate the loss with the rendered color $\hat{c}$ which becomes the main loss over time. This loss is computed over multiple resolution levels as outlined in Sec.~\ref{sec:method} of the main paper whenever patches are rendered.

\inlinesection{Mask losses.} %
In total we use three mask loss terms. The $\loss{silhouette}$ as described in Sec~\ref{sec:losses_optimization} as well as the binary cross-entropy loss $\loss{BCE}$ between the volume-rendered mask and estimated foreground object mask and the background loss $\loss{Background}$ from NeRD~\cite{Boss2021}. The latter enforces all rays cast to the background to return $0$. Consequently, the total mask loss is defined as: $\loss{Mask} = \lambda_\mathrm{xor}\loss{silhouette} + \loss{BCE} + \loss{Background}$ where $\lambda_\mathrm{xor}$ is set to $50$ and $\loss{silhouette}$ is normalized by the number of elements in the reference mask.

\inlinesection{Final loss ensemble.}
Overall we compute two loss terms $\loss{Network}$ and $\loss{Camera}$ which consist of differently weighted versions of the photometric rendering loss and alignment losses plus the respective regularizations.
The loss to optimize the decomposition network can be written as $\loss{Network}=\lambda_b \loss{Image}(\vect{C^s}, \vect{\tilde{c}}) + (1 - \lambda_b) \loss{Image}(\vect{C^s}, \vect{\hat{c}}) + \loss{Mask} + \lambda_a \loss{Init} + \lambda_\mathrm{ndir} \loss{ndir} + \lambda_\mathrm{Smooth} \loss{Smooth} + \lambda_\mathrm{Dec Smooth} \loss{Dec Smooth} + \lambda_\mathrm{Dec Sparsity} \loss{Dec Sparsity}$. Here, $\lambda_b$ and $\lambda_a$ are the optimization scheduling weights described below in more detail. As long as the camera multiplex has size $m>1$ the camera multiplex consistency loss is added as follows: $\loss{Network} = \loss{Network} + 0.1 (\loss{multiplex})$.
To these losses the camera posterior scaling is applied as in SAMURAI~\cite{bossSAMURAIShapeMaterial2022}.
The camera loss is weighted according to our view importance scaling instead. 
Badly initialized camera poses can still recover over the training duration as they get potentially large updates while cameras that perform well in terms of the losses are gradually faded out of the optimization. Additionally, the regularizations from above, $\loss{Bounds}$ and $\loss{Lookat}$ are added.

\begin{figure}[ht]
\begin{minipage}{\linewidth}

         \includegraphics[width=\textwidth]{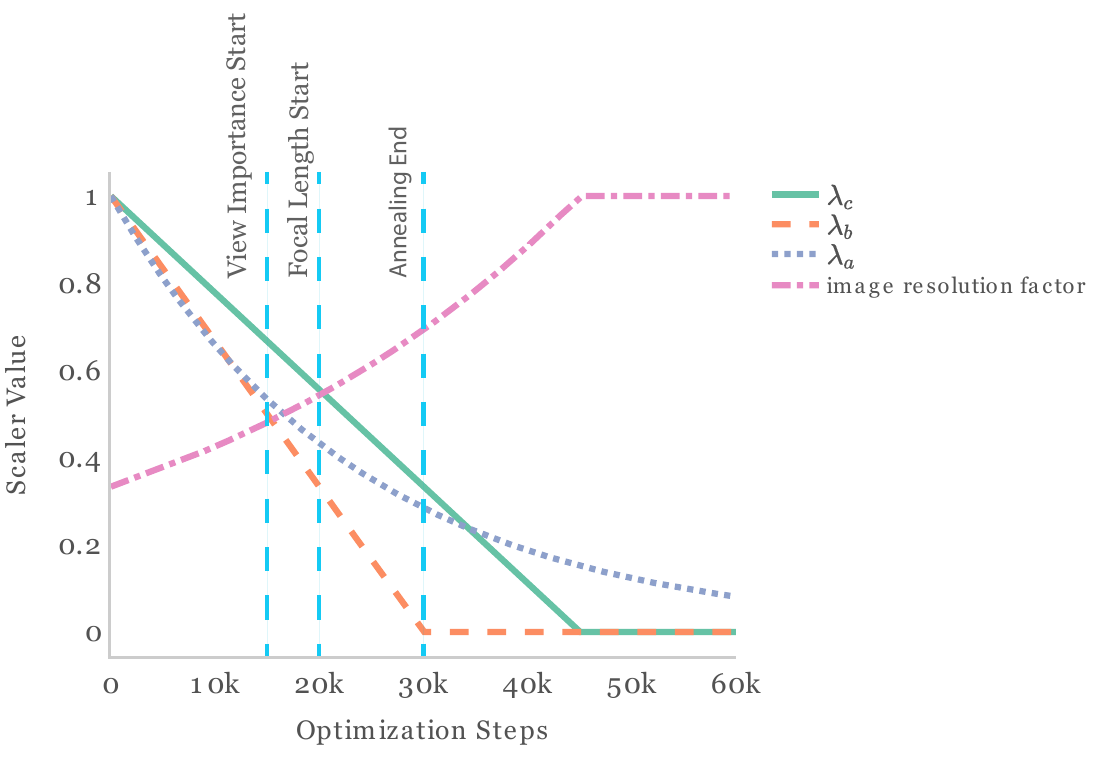}
\caption{\small \textbf{Optimization schedule.} We use three $\lambda$ parameter to scale losses to enable a smooth flow of the optimization parameters. Additionally, we indicate at which points in time the view importance weighting is introduced, the focal length parameters start to get updated and the encoding annealing ends.}
\label{fig:sup_optimization}
\end{minipage}
\end{figure}

\subsection{Optimization}
\label{sec:sup_method_details_optimization}
\inlinesection{Optimization scheduling.}
We use three fading $\lambda$ variables to steer the optimization schedule smoothly as visualized in \fig{sup_optimization}. Render resolution is continuously increased over the first half of the optimization while the number of active multiplex cameras is reduced. This is controlled by $\lambda_c$. Input image resolution is increased from 100 pixels to a resolution of 400 pixels on the longer image side over the first half of the training. For higher final output resolutions an even larger downsample factor ($>4$) might be needed. This strategy allows the image patches to include even larger structures of the objects and improves camera alignment. The direct color optimization is faded to the BRDF optimization and the encoding annealing is performed over the first third of the optimization. $\lambda_b$ is used for the BRDF transition and an independent $\alpha$ value is kept for the annealing. Finally, $\lambda_a$ is used to scale some losses in a non-linear way. Focal length updates are delayed until a quarter of the optimization time. We start with the view importance weighting at the half-way point of the annealing schedule.
SHINOBI renders image patches for most of the training time which adds more context to each update step, allowing us to add new losses tailored to camera alignment. The first 1000 steps are trained using regular random ray sampling, though, to help initialize a global shape quickly while both the render resolution as well as the hash grid resolution are low.

\inlinesection{Optimizer settings.}
The ADAM~\cite{kingma2014adam} optimizer updates the network weights based on $\loss{Network}$ with a learning rate of \expnumber{1}{-3} that is exponentially decayed by an order of magnitude over the training time. The same decay rate is applied to the optimizer concerned with the hash grid embeddings. The gradient are computed based on $\loss{Network}$ with the hash grid specific regularization $\loss{Grid}$ added. The learning rate of the camera optimizer is exponentially decayed by an order of magnitude every 40k steps. As mentioned before the $\beta1$ parameter is set to $0.2$ for the camera optimizer to stabilize the training in the presence of noisy gradients. It uses the gradients computed based on $\loss{Camera}$.
The framework is trained using float16 mixed precision. The coordinate input to the encoding is $32$ bit as is the rendering and illumination evaluation. The other MLPs and specifically the interpolation on the hash grids run at $16$ bit precision, though. 

\section{Additional Experiments}
\label{sec:sup_experiments}

\subsection{Details on Compared Methods.}
\label{sec:sup_exp_baselines}
In addition to SAMURAI which has been introduced in Sec.~\ref{sec:method} of the main paper we compare against two more recent methods for in-the-wild object reconstruction.

\inlinesection{NeRS}
\label{sec:sup_exp_baselines_ners}
stands for Neural Reflectance Surfaces~\cite{zhang2021ners} that constrain reconstructions using a mesh-based representation. Starting from manually annotated rough initial poses and a template mesh the objects are decomposed into a surface mesh, illumination and surface reflectivity parameterized as albedo and shininess. We define the dimensions of an initial cuboid that approximates the object's bounding box for each scene in line with \cite{zhang2021ners, jampani2023navi}. 

\inlinesection{NeROIC}
\label{sec:sup_exp_baselines_neroic}
presents a multi-stage approach to reconstruct geometry and material properties of objects from online image collections. Camera poses are initialized with a COLMAP-based pipeline and fine-tuned during the first reconstruction stage. Following high-quality surface normals are estimated during the second stage. Finally, material properties and illumination are optimized to enable relighting in addition to novel view synthesis.

\subsection{Additional Visual Results}
\label{sec:sup_exp_navi}
\fig{sup_comparison_view_synthesis} shows additional qualitative results on objects from the NAVI dataset compared to the baseline methods. Note, that the methods work at different image resolutions and that we show the original output. NeROIC is able to reconstruct high-frequency detail for scenes that have good initial poses but shows artifacts or fails on others. NeRS suffers from its low resolution mesh representation and often inaccurate camera alignment.
SAMURAI and SHINOBI both reproduce appearance that is closer to the original illumination setting due to superior decomposition capabilities while SHINOBI recovers more high-frequency details. Consequently, on the reference free sharpness metric CPBD~\cite{Narvekar2011CPBD} our method clearly improves upon SAMURAI and NeRS, with CPBD scores of \colorbox{bestcol}{\textbf{0.82}} (Ours) vs. \colorbox{secondbestcol}{0.77} for SAMURAI vs. 0.65 for NeRS on the most challenging subset of the NAVI scenes where COLMAP reconstruction fails.

\begin{figure}[ht]
\begin{minipage}{\linewidth}

         \includegraphics[width=\textwidth]{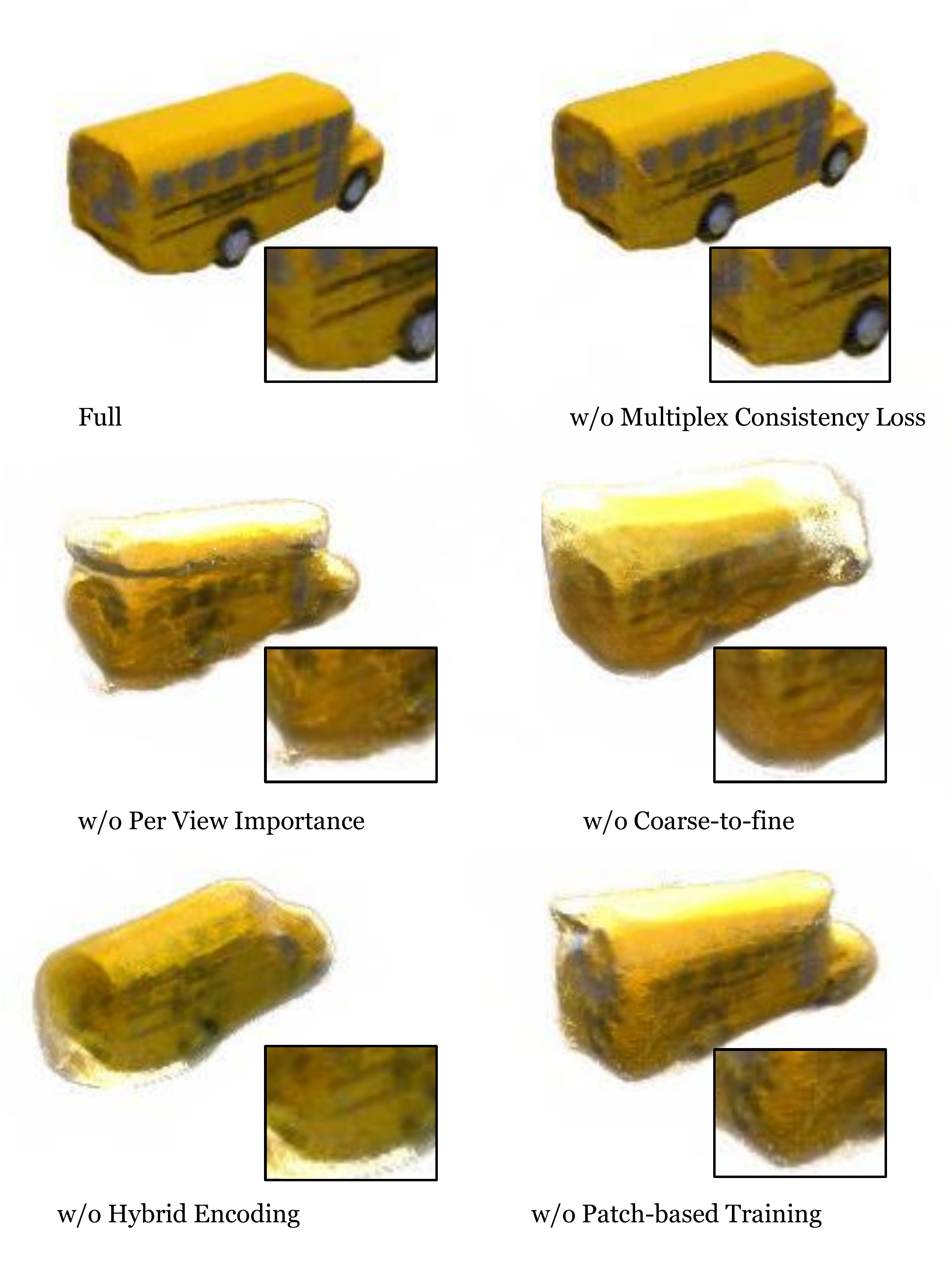}
\caption{\small \textbf{Qualitative ablation study.} We show view synthesis results from novel view synthesis on the `School Bus' scene from NAVI where we ablate components of our method. The visual results underline the importance of each part.}
\label{fig:sup_ablations}
\end{minipage}
\end{figure}

\begin{figure}
\begin{minipage}{\linewidth}
    \centering
     \begin{subfigure}[b]{0.4\textwidth}
         \centering
         \includegraphics[trim={0 1cm 0 0.3cm},clip, width=\textwidth]{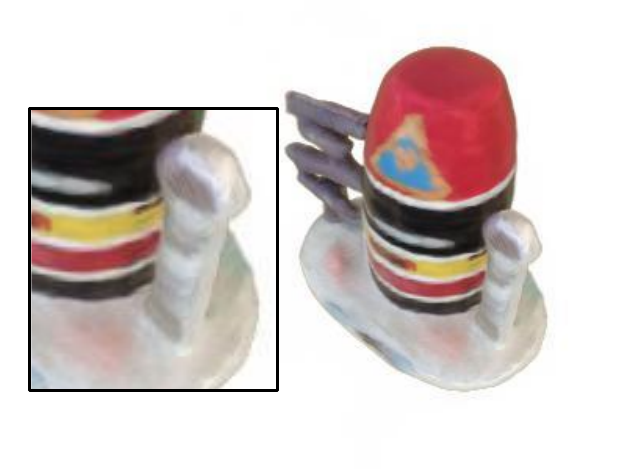}
         \caption{\footnotesize Only Fourier}
     \end{subfigure}
     \begin{subfigure}[b]{0.4\textwidth}
         \centering
         \includegraphics[trim={0 1cm 0 0.3cm},clip,width=\textwidth]{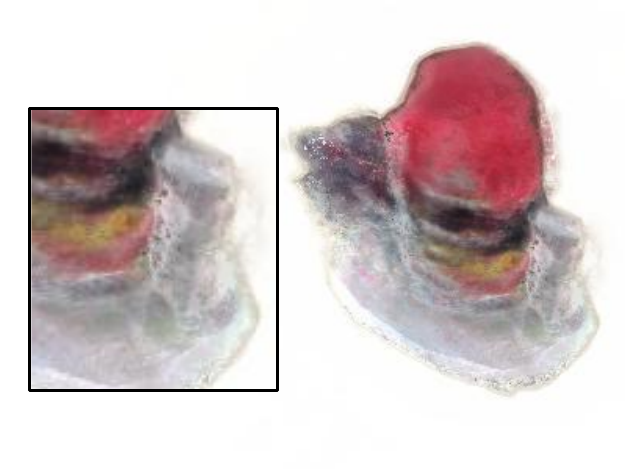}
         \caption{\footnotesize Fourier faded out}
     \end{subfigure}
     \hfill
     \\
     \begin{subfigure}[b]{0.4\textwidth}
         \centering
         \includegraphics[trim={0cm 1cm 0.3cm 0.3cm},clip,width=\textwidth]{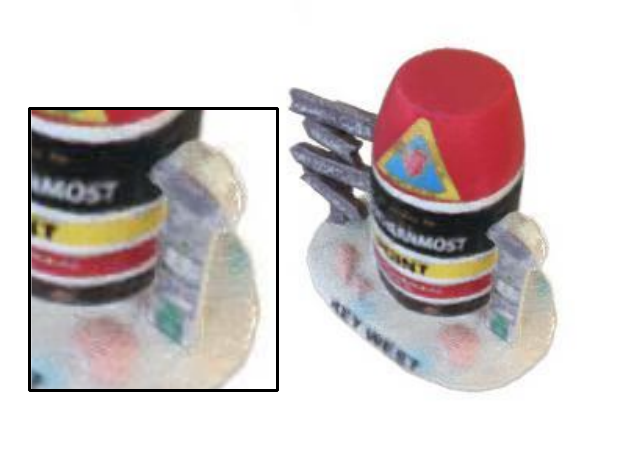}
         \caption{\footnotesize Eval hash grid only}
     \end{subfigure}
     \begin{subfigure}[b]{0.4\textwidth}
         \centering
         \includegraphics[trim={0 1cm 0 0.3cm},clip,width=\textwidth]{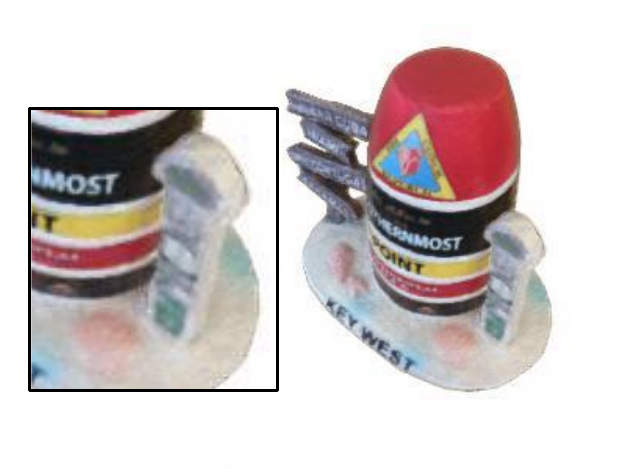}
         \caption{\footnotesize Hybrid encoding}
     \end{subfigure}
     \vspace{-2mm}
        \caption{\footnotesize \textbf{Hybrid Encoding Ablation.} The Fourier encoding on its own is band limited, e.g. text is not reconstructed. Fading out the Fourier encoding during training destabilizes the optimization showing the importance of both encoding schemes. Evaluation using only the hash grid encoding results in noisy density but sharp texture. Our hybrid encoding yields sharp results with a consistent density at the object's surface.}
        \label{fig:reb_hybrid_encoding}
\vspace{-4mm}
\end{minipage}
\end{figure}

\subsection{Qualitative Results of Ablations}
\label{sec:sup_exp_ablations}
\fig{sup_ablations} shows qualitative results corresponding to the numerical results from the ablation study reported in the main paper. It can be observed that a robust reconstruction is only possible using the full configuration of our method. While the multiplex consistency loss has only minimal impact on this example the result still shows some visible artifacts and overall increased noise level.
It is apparent that a plain integration of multi-resolution hash grid encoding does not perform well on the task of joint camera pose and shape reconstruction
Fig.~\ref{fig:reb_hybrid_encoding} visualizes how the encoding schemes complement each other. It has been shown that Fourier encoding together with a large MLP and a coarse-to-fine scheme does perform reasonable well for camera optimization~\cite{bossSAMURAIShapeMaterial2022, lin2021barf}. Hence, we merge the two encoders while keeping the MLP small and therefore band-limited. The main advantage is the continuity of the gradients for the coarse geometry that help to pull the optimization targets closer to the optimum in the beginning of the optimization. Updates based on higher-frequency details later in the optimization can propagate through the hash grids as they are constrained by our additional losses.

\vspace{2mm}

\begin{table}[ht]
\resizebox{\linewidth}{!}{ %
\Huge
\begin{tabular}{@{}
l
S[table-format=1.2(3)]
S[table-format=1.2(3)]
S[table-format=3.2(4)]
S[table-format=3.2(4)]
@{}}
\multicolumn{1}{c}{Method} 
& \multicolumn{2}{c}{Translation\textdownarrow} 
& \multicolumn{2}{c}{Rotation \textdegree \textdownarrow}
\\
\cmidrule(lr){2-3}
\cmidrule(lr){4-5}
 & 
 {$S_C$} & {$\sim S_C$} & {$S_C$} & {$\sim S_C$}
\\
\midrule
PoseDiffusion~\cite{wang2023posediffusion} & 0.51\pm 0.09 & 0.43\pm 0.11 & 
41.33\pm 15.15 & 43.50\pm 13.67
\\
HLoc~\cite{sarlin2019hloc,sarlin2020superglue} & 0.07\pm 0.13 & 0.06\pm 0.10 & 9.1\pm 18.75& 9.72\pm 20.08
\\
\OURS  &
 0.25\pm 0.085 & 0.28\pm 0.09 & 22.84\pm 16.19 & 33.00\pm 19.97
\\
\end{tabular}
} %
\begingroup %
\titlecaptionof{table}{\small Pose estimation on in-the-wild data.}{\small Evaluation of absolute rotation and translation errors after alignment on the NAVI~\cite{jampani2023navi} in-the-wild scenes. We compare~\OURS against specialized camera pose estimation solutions. Note, that HLoc~\cite{sarlin2019hloc} fails completely on 5 scenes and is only able to recover 55$\%$ of views on average.}
\label{tab:sup_pose_estimation}
\endgroup
\end{table}

\subsection{Comparison to other Camera Pose Estimation Methods}
Tab.~\ref{tab:sup_pose_estimation} compares methods for camera pose estimation on the NAVI in-the-wild scenes~\cite{jampani2023navi}. Traditional SfM methods like COLMAP~\cite{schoenberger2016mvs,schoenberger2016sfm} paired with a neural feature detection and matching can recover poses with great accuracy but only succeed on a subset of scenes and images. PoseDiffusion~\cite{wang2023posediffusion} and ID-Pose~\cite{cheng2023idpose}, both fully neural models trained on large datasets, struggle on these out-of-distribution examples. We only report a full evaluation on PoseDiffusion as an example here. We observe that these models take important pose cues also from the background of object-centric image sets. This leads to poor results on in-the-wild image collections. A simple fine-tuning on masked images did not improve performance. In our experiments, camera pose estimation usually regresses to a front-facing camera layout for in-the-wild examples featuring different illumination and object scales. Consequently, our approach appears to be a good trade-off in-terms of camera pose quality.
\begin{figure}
\begin{minipage}{\linewidth}
    \centering
     \begin{subfigure}[b]{\textwidth}
         \centering
         \includegraphics[width=\textwidth]{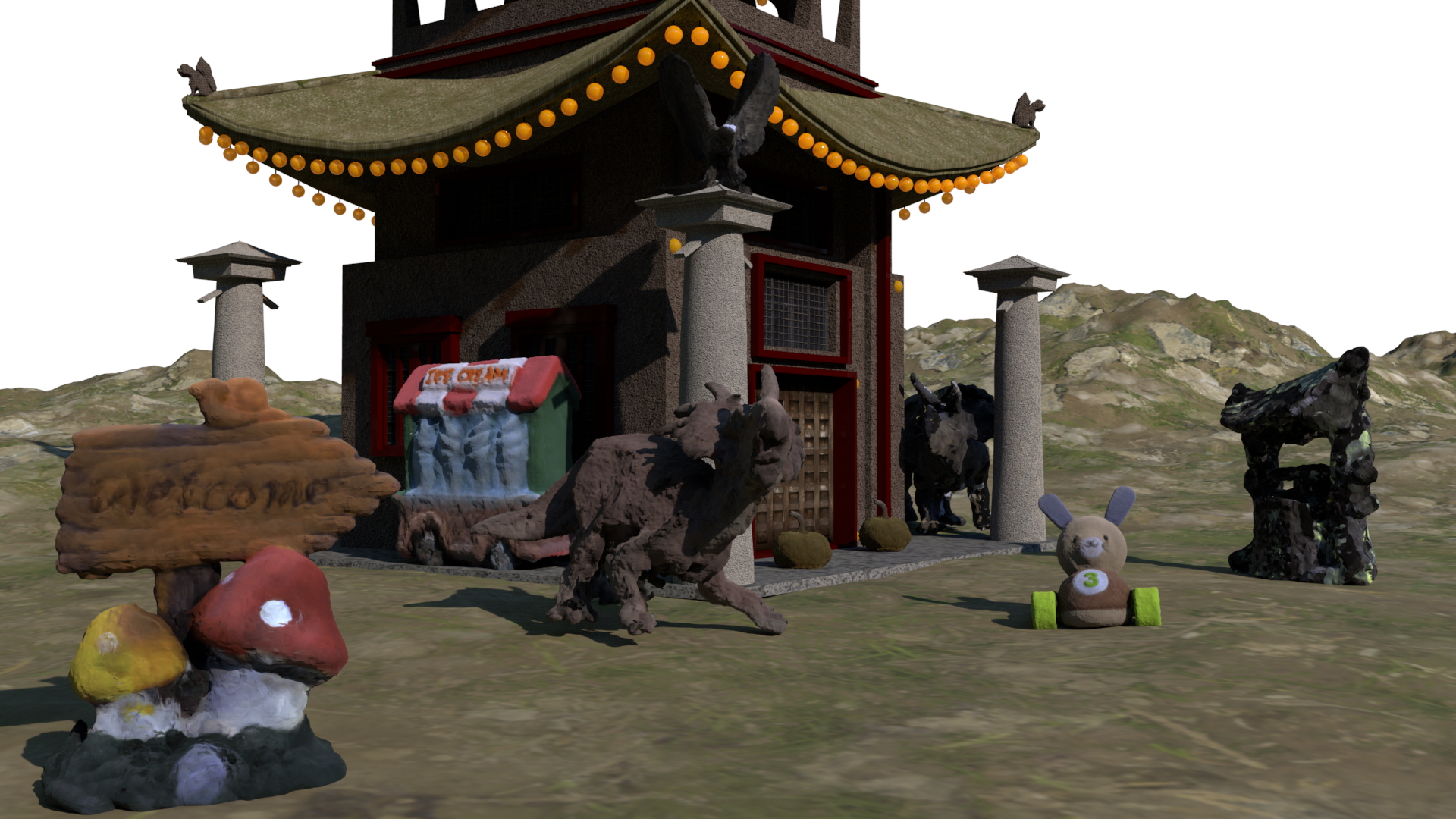}
         \caption{\footnotesize Reconstructed assets under novel illumination}
     \end{subfigure}
     \\
     \begin{subfigure}[b]{\textwidth}
         \centering
         \includegraphics[width=\textwidth]{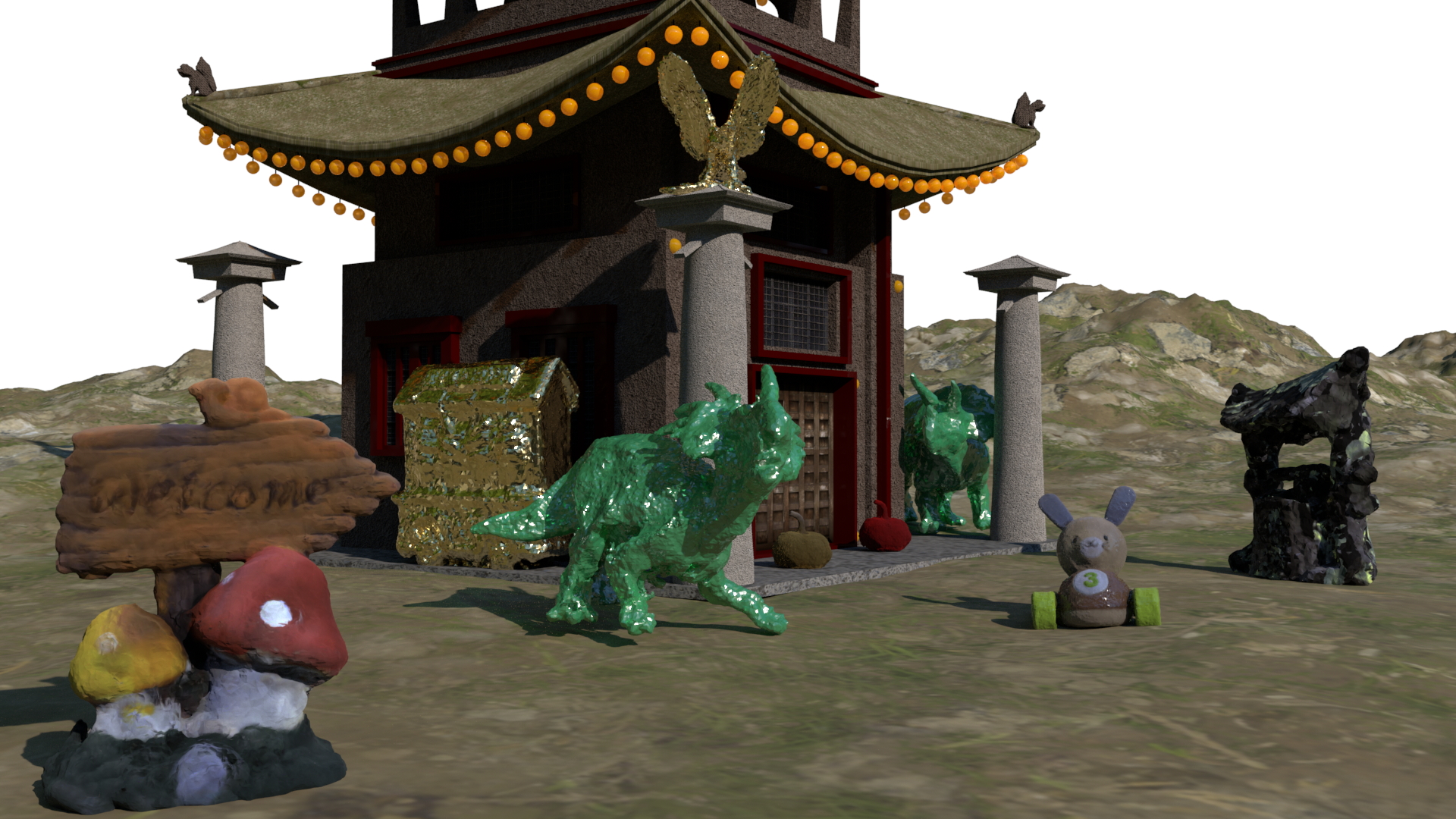}
         \caption{\footnotesize Edited materials}
     \end{subfigure}
     \vspace{-2mm}
        \caption{\small \textbf{Integration and editing.} Although objects are initially captured under diverse illumination settings we can  integrate multiple objects consistently into a scene in the end. BRDF parameters can be modified independently from the illumination.}
        \label{fig:applications}
\vspace{-4mm}
\end{minipage}
\end{figure}

\begin{figure}
\begin{minipage}{\linewidth}
    \centering
     \begin{subfigure}[b]{0.32\textwidth}
         \centering
         \includegraphics[width=\textwidth]{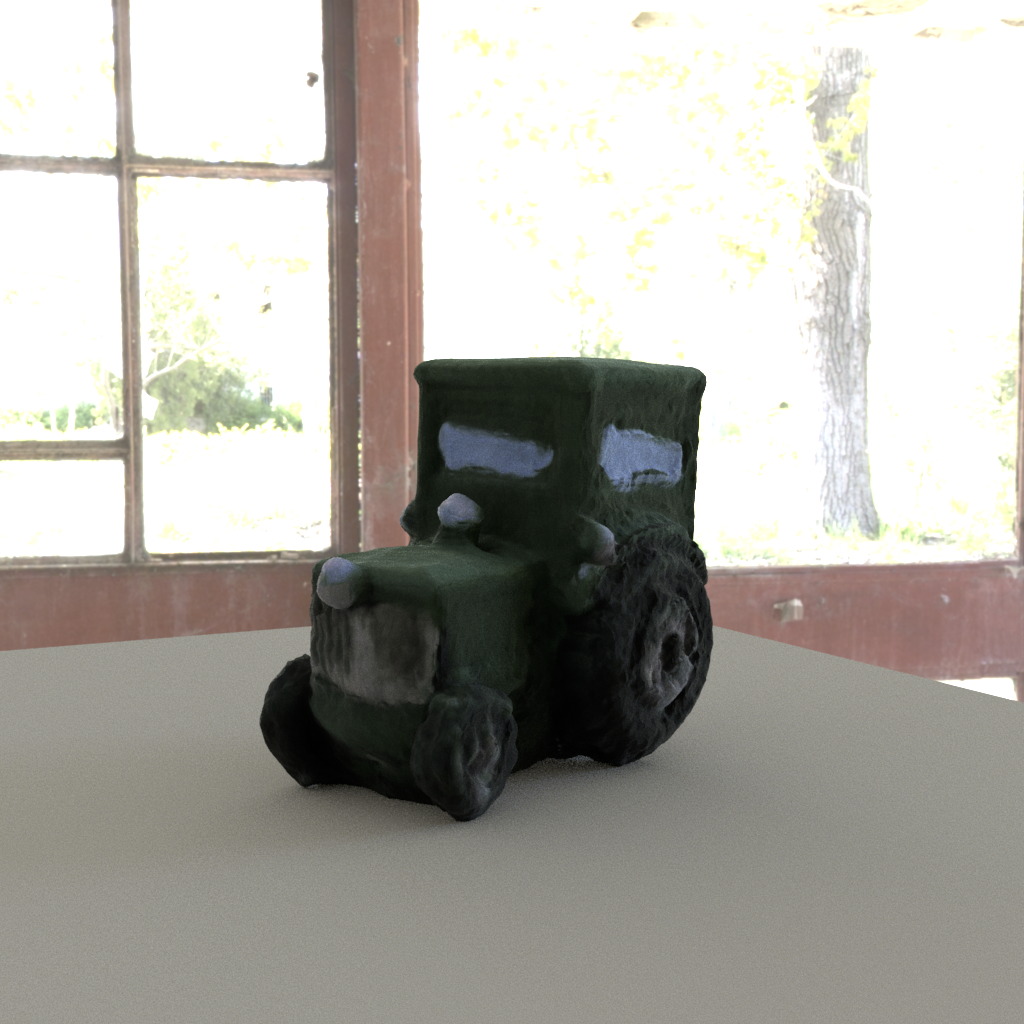}
     \end{subfigure}
     \hfill
     \begin{subfigure}[b]{0.32\textwidth}
         \centering
         \includegraphics[width=\textwidth]{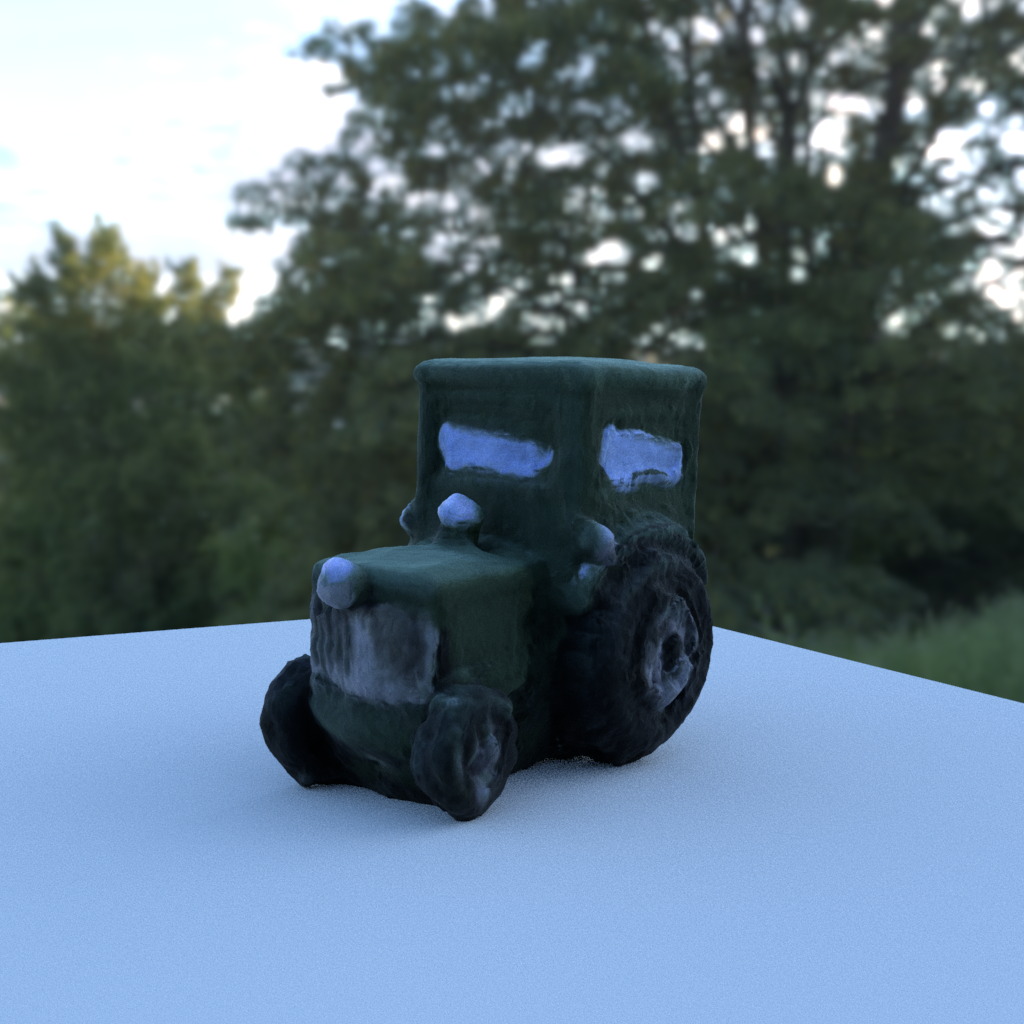}
     \end{subfigure}
     \hfill
     \begin{subfigure}[b]{0.32\textwidth}
         \centering
         \includegraphics[width=\textwidth]{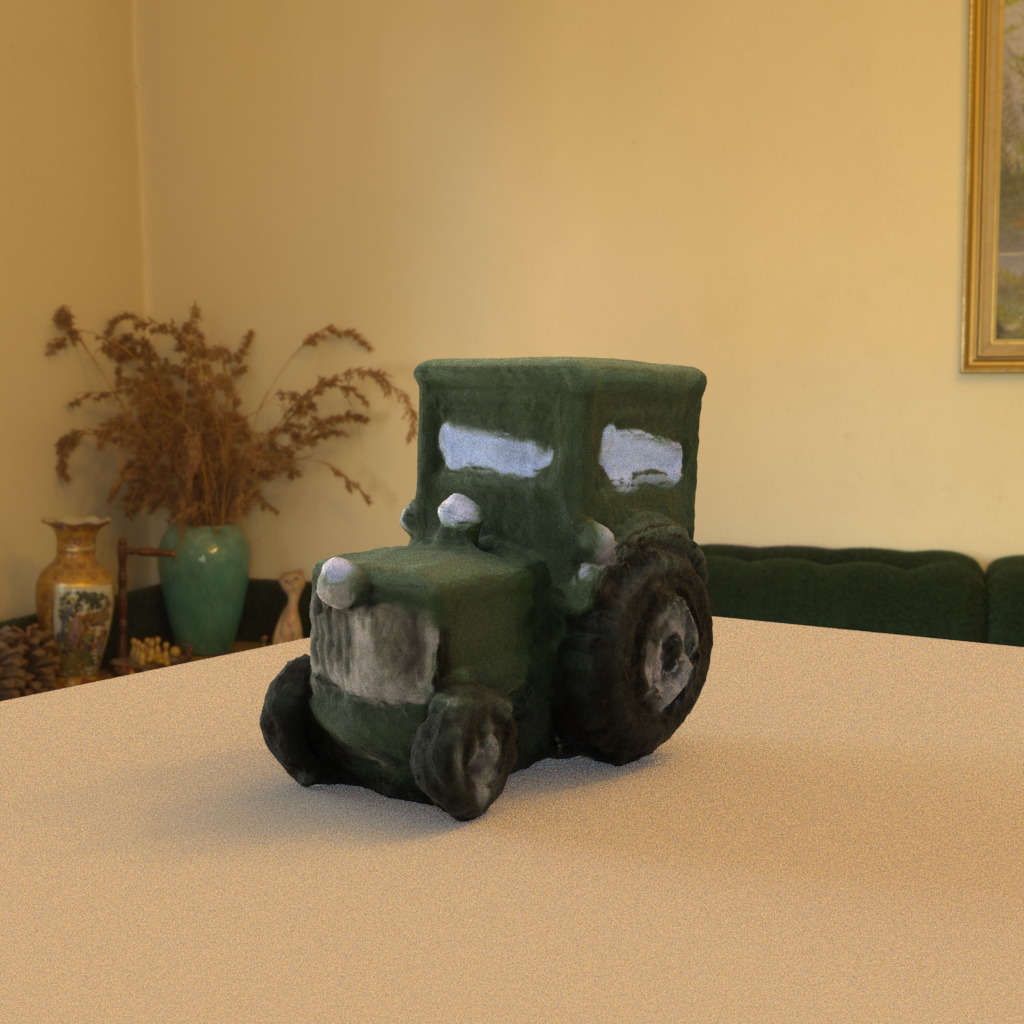}
     \end{subfigure}
     \vspace{-2mm}
        \caption{\small \textbf{Relighting application.} View synthesis under three different illumination settings using the estimated decomposition for a sample view from the ``Tractor" scene.}
        \label{fig:relighting}
\vspace{-4mm}
\end{minipage}
\end{figure}

\subsection{Downstream Applications}
\label{sec:sup_exp_applications}
The object decomposition into BRDF, illumination and shape enables us to edit illumination and material independently of the shape representation to re-light the object, for example. Furthermore, we can convert our neural representation into a parametric model like a mesh and physically based material suitable for easy integration into standard graphics pipelines.
\inlinesection{Mesh extraction and asset generation.}
We use a modified version of the mesh extraction component from SAMURAI~\cite{bossSAMURAIShapeMaterial2022} to extract triangle meshes from the learnt volume and the corresponding material parameters. 
Marching cubes is used to create an initial mesh. We post-process the mesh and perform automatic UV unwrapping using Blender~\cite{blender}. Finally, textures are extracted by querying our pipeline for the BRDF around the baked surface locations. The extraction of a mesh takes around 3 minutes.

\inlinesection{Relighting and material editing.}
Our reconstructed assets can then be easily integrated into existing graphics pipelines. In \fig{applications} we show a SHINOBI themed scene featuring objects from the NAVI dataset in a new consistent illumination environment as it would be required for AR and VR applications. We can also modify the BRDF parameters independently of the lighting. \fig{relighting} compares renderings of the same camera view but lit with different environment lights. Please also consider watching the supplementary video including more examples for the given applications.

 \fi

\end{document}